\documentclass{article}

\usepackage{stylefile}
\usepackage[numbers]{natbib}

\usepackage[T1]{fontenc}                    
\usepackage[utf8]{inputenc}                  
\usepackage{hyperref}                       
\usepackage{url}                            
\usepackage{booktabs}                     
\usepackage{amsfonts}                        
\usepackage{nicefrac}                      
\usepackage{microtype}                      
\usepackage{amsmath}                        
\usepackage{amsthm}                         
\usepackage{cleveref}                       
\usepackage{graphicx}                       
\usepackage{wrapfig}                       
\usepackage{algorithm}                      
\usepackage{algorithmic}                    
\usepackage[algo2e]{algorithm2e}                    
\usepackage{subcaption}                     
\usepackage{lipsum}                         
\usepackage[explicit]{titlesec}              
\usepackage{multirow}                       
\usepackage{xcolor}

\titlespacing{\paragraph}{0pt}{0pt}{.2em}[]
\definecolor{regcolor}{HTML}{1F77B4}
\definecolor{starcolor}{HTML}{ff7f0e}

\newcommand{\eg}{\textit{e.g.,} }
\newcommand{\ie}{\textit{i.e.,} }
\newcommand{\quotes}[1]{``#1''}
\newcommand{\viz}{\textit{viz.,} }

\newcommand{\ourmethod}{Starlight\xspace}
\newcommand{\regcolor}[1]{\textcolor{regcolor}{#1}}
\newcommand{\starcolor}[1]{\textcolor{starcolor}{#1}}

\newcommand{\genericloss}[1]{\mathcal{L}\left(#1\right)}
\newcommand{\genericlosswithdataset}[1]{\mathcal{L}\left(#1; \dataset\right)}
\newcommand{\dataset}{\mathcal{D}}
\newcommand{\winperm}[2]{\underset{#1\rightarrow#2}{\pi}}
\newcommand{\allperms}[1]{\mathcal{P}_{#1}}
\newcommand{\starmodel}[1]{\theta^\star_{#1}}
\newcommand{\barrier}[2]{B\left(#1, #2\right)}
\newcommand{\sdloss}[2]{\widetilde{\mathcal{L}}_{#2}(#1)}
\newcommand{\solutionset}{S}
\newcommand{\heldout}{H}
\newcommand{\sourcemodels}{Z}
\DeclareMathOperator*{\argmin}{arg\,min}

\theoremstyle{plain}

\newtheorem{conjecture}{Conjecture}
\theoremstyle{definition}

\theoremstyle{remark}

\crefname{figure}{Figure}{Figures}
\Crefname{figure}{Figure}{Figures}
\crefname{conjecture}{Conjecture}{Conjectures}
\Crefname{conjecture}{Conjecture}{Conjectures}
\crefname{algorithm}{Algorithm}{Algorithms}
\Crefname{algorithm}{Algorithm}{Algorithms}
\crefname{appendix}{Appendix}{Appendix}
\Crefname{appendix}{Appendix}{Appendix}
\crefname{table}{Table}{Tables}
\Crefname{table}{Table}{Tables}
\crefname{section}{Section}{Sections}
\Crefname{section}{Section}{Sections}

\title{Do Deep Neural Network Solutions Form a Star Domain?}

\author{
  Ankit Sonthalia \\
  T{\"u}bingen AI Center \\
  Universit{\"a}t Tübingen, Germany \\
  \texttt{ankit.sonthalia@uni-tuebingen.de} \\
  \And
  Alexander Rubinstein \\
  T{\"u}bingen AI Center \\
  Universit{\"a}t Tübingen, Germany \\
  \And
  Ehsan Abbasnejad \\
  Australian Institute for Machine Learning \\
  University of Adelaide, Australia \\
  \And
  Seong Joon Oh \\
  T{\"u}bingen AI Center \\
  Universit{\"a}t Tübingen, Germany \\
}

\begin{document}

\maketitle

\begin{abstract}
  It has recently been conjectured that neural network solution sets reachable via stochastic gradient descent (SGD) are convex, considering permutation invariances \cite{entezari_permutation_invariances_2022}. This means that a linear path can connect two independent solutions with low loss, given the weights of one of the models are appropriately permuted. However, current methods to test this theory often require very wide networks to succeed \citep{ainsworth_git_re_basin_2022, benzing_random_2022}. In this work, we conjecture that more generally, the SGD solution set is a \textit{star domain} that contains a \textit{star model} that is linearly connected to all the other solutions via paths with low loss values, modulo permutations. We propose the \textit{Starlight} algorithm that finds a star model of a given learning task. We validate our claim by showing that this star model is linearly connected with other independently found solutions. As an additional benefit of our study, we demonstrate better uncertainty estimates on the Bayesian Model Averaging over the obtained star domain. Further, we demonstrate star models as potential substitutes for model ensembles. Our code is available at \href{https://github.com/aktsonthalia/starlight}{https://github.com/aktsonthalia/starlight}.
\end{abstract}

\section{Introduction}

The learning problem for a neural network is inherently 
characterized by a non-convex loss landscape, leading to multiple possible solutions rather than a singular one. Efforts to comprehend this landscape and the set of solutions have been ongoing.

A significant early discovery in this area \cite{garipov_mode_connectivity_2018} demonstrated that almost any two independent solutions could be connected through a simple low-loss curve. While this finding highlighted the vastness of the solution set, other research has focused on its complexity. 

For instance, permutation symmetries allow neuron positions in different layers to be jointly swapped without changing the function represented by the neural network \citep{brea2019weight, singh_jaggi_otfusion_2021, ainsworth_git_re_basin_2022, sinkhorn_re_basin_2023}. \cite{entezari_permutation_invariances_2022} proposed that when accounting for these symmetries, the solution set found by stochastic gradient descent (SGD) essentially becomes convex, \ie any pair of independent solutions can be connected through a low-loss line segment after an appropriate permutation is applied to one of the models. Notably, \cite{sharma2024simultaneous} investigate the stronger property of \emph{simultaneous} linear connectivity, wherein permuting a given model linearly connects it to \textit{several} other models. However, recent works \citep{ainsworth_git_re_basin_2022, xiao2023compact} study convexity in the context of the formulation in \cite{entezari_permutation_invariances_2022}. Our work therefore refers to their conjecture as the \quotes{convexity conjecture} (\cref{conjecture:convexity}) while acknowledging that other, stronger forms of convexity can be formulated.

The convexity conjecture has faced challenges.
Subsequent studies \citep{juneja2022linearreveals, benzing_random_2022, ainsworth_git_re_basin_2022, altintas_disentangling_2023, sinkhorn_re_basin_2023} revealed that even after the application of permutation-finding algorithms,
two distinct solutions in the parameter space might still be separated by a high loss barrier \citep{frankle2020lottery, entezari_permutation_invariances_2022} upon performing linear interpolation. 
These studies attribute this discrepancy to various factors, including network depth and width, dataset complexity \citep{ainsworth_git_re_basin_2022} and high learning rates \citep{altintas_disentangling_2023}. Theoretical investigation \citep{entezari_permutation_invariances_2022, ferbach2024proving} suggests that in general, the conjecture needs wide networks to hold, and that deeper networks might need to be even wider than their shallower counterparts to satisfy the conjecture.

\begin{wrapfigure}{r}{0.7\linewidth}
\includegraphics[width=\linewidth]{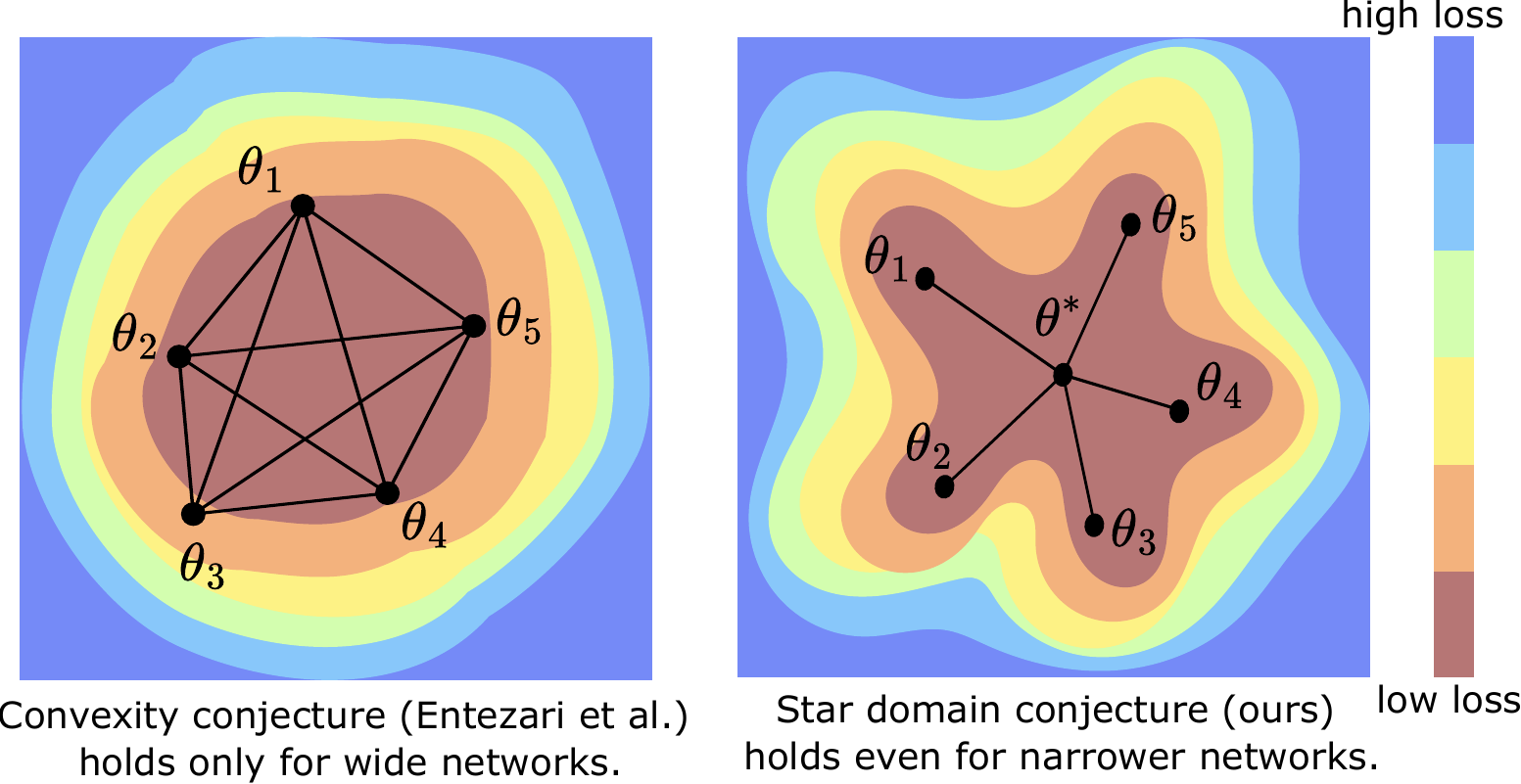}
\label{figure:posterfigure}
\vspace{-1em}
\end{wrapfigure}

In response to these findings, our research introduces the \textbf{star domain conjecture}. 
We propose that solutions in deep neural networks (DNNs) form a star domain rather than a convex set, modulo permutation symmetries. 
A \textit{star domain} is a set $A$ with at least one special element, known as a \textit{star point}, $a_0\in A$ that is connected to every other element in $A$. 
A convex set is a specific instance of a star domain. 
The star domain conjecture thus proposes that in cases where convexity \citep{entezari_permutation_invariances_2022} does not hold, a weaker form of convexity (\ie star-shaped connectivity) still exists.  

The star domain conjecture is still a stronger assertion than mode connectivity \citep{garipov_mode_connectivity_2018} which states that any two models $\theta_A$ and $\theta_B$ can be connected through a possibly non-linear path in the solution space. As a special case, this path could be as simple as a piece-wise linear path comprising a third point $\theta_C$ such that $(\theta_A,\theta_C)$ and $(\theta_B,\theta_C)$ are linearly connected. In contrast, our conjecture implies that \textit{all} pairs of solutions are interconnected via a \textit{shared} third solution, the star point, which is \textit{common} to all solution pairs: $\exists \theta_C$ such that $\forall \theta_A,\theta_B\in\solutionset$, $(\theta_A,\theta_C)$ and $(\theta_B,\theta_C)$ are linearly connected, where $\solutionset$ is the solution set.

We substantiate our star domain conjecture with empirical evidence by introducing the \ourmethod algorithm to identify a candidate star model for a given learning task. \ourmethod finds a model that is linearly connected with a finite set of independent solutions. We demonstrate that these star model candidates have low loss barriers with an arbitrary set of solutions that were not used in constructing the star model candidates. This provides strong evidence that there exist star models that are linearly connected with other solutions.

In addition to validating the conjecture, our research delves into the distinctive characteristics of star models. We find that sampling from the star domain for Bayesian Model Averaging (BMA) leads to better uncertainty estimates than ensembles. Additionally, we demonstrate star models as a possible substitute to model ensembles, with lower inference time and memory footprint. These differences highlight the potential advantages of star models in various neural network applications.

We summarise our contributions:

\begin{enumerate}
    \item The \textbf{star domain conjecture} for characterizing connectivity in neural network solution sets.
    \item The \textbf{\ourmethod} algorithm for identifying a star model for a gradient-based learning task.
    \item Analysis of practical benefits shown by the star models.
\end{enumerate}

\section{Related work}
\label{section:related_work}

We introduce the relevant development of findings toward the understanding of DNN solution sets.

\textbf{Mode Connectivity.} 
\cite{garipov_mode_connectivity_2018} and \cite{draxler_essentially_no_barriers2018} concurrently discovered mode connectivity. 
\cite{gotmare2018using_mode_connectivity} soon followed, showing non-linear connectivity even between networks obtained using different training schemes. \cite{kuditipudi_explaining_connectivity_dropout_noise_2019} explained mode connectivity via dropout stability and noise stability. 
\cite{benton_loss_simplexes_2021} went on to show that there exist not only simple paths, but also \emph{volumes} of low loss, connecting several DNN solutions. These works focus on general, \emph{non-linear} connectivity, while we study a stricter condition, \viz linear connectivity. 

\textbf{Linear Mode Connectivity (LMC).} \cite{frankle2020lottery} were the first to study LMC. Later, \cite{entezari_permutation_invariances_2022} proposed that SGD solution sets are convex modulo permutations, while \cite{singh_jaggi_otfusion_2021, ainsworth_git_re_basin_2022, sinkhorn_re_basin_2023} introduced "re-basin" methods, \ie methods for bringing different solutions into the same basin. Recent work \citep{ainsworth_git_re_basin_2022, altintas_disentangling_2023, benzing_random_2022} also noted failure cases for LMC, while
\cite{ferbach2024proving} theoretically investigated convexity for sufficiently wide nets.
Our analysis  builds upon these findings and reveals evidence for a weaker property, \viz star-shaped connectivity, in cases where convexity does not hold.

\textbf{Star-shaped connectivity (SSC).} \cite{annesi2023star} provide valuable insights for SSC in the loss landscape for the simple case of the negative spherical perceptron. In contrast, we consider more complex models and learning tasks, and propose a novel verification method for SSC. 
Concurrently to us, \cite{lin2024exploring} obtain star models linearly connected with a \textit{finite} number of solutions to larger nets like the VGG16 \citep{simonyan2014vgg}.
In contrast, our work additionally considers permutation invariances \citep{ainsworth_git_re_basin_2022, entezari_permutation_invariances_2022} and provides evidence that star models trained this way might be connected to \textit{infinitely many} other solutions. 

\textbf{Practical Applications.} Mode connectivity has found applications in
model fusion (\cite{garipov_mode_connectivity_2018, singh_jaggi_otfusion_2021}), adversarial robustness (\cite{zhao_adversarial_bridging_mode_connect_2020, wang_robust_mode_connectivity_2023}),
continual learning (\cite{mirzadeh_lmc_multitask_contiunal_2020, wen_opc_connectivity_2023}), and federated learning (\cite{wang2020federated, ainsworth_git_re_basin_2022}). In contrast, our work focuses on understanding the surface of the loss landscape. However, we also explore potential applications, \eg Bayesian Model Averaging.

\section{The star domain conjecture}

\subsection{Background: the convexity conjecture}
\label{section:background_convexity}

Here, we formally state the convexity conjecture, starting with basic notations. A neural network is a function $f_{\theta}(\cdot)$ parameterized by $\theta \in \Theta$, where $\Theta$ is the parameter space.
Given a dataset $\dataset$, we formulate a non-negative loss $\mathcal{L}(\theta)=\genericlosswithdataset{\theta}\geq 0$ and minimize $\genericloss{\theta}$ to find a solution in $\Theta$.

The \textbf{solution set} is $S:=\{\theta \mid \genericloss{\theta} \approx 0\}$.  

The \textbf{loss barrier} was first defined by \cite{frankle2020lottery}. We use the formulation in \cite{entezari_permutation_invariances_2022}, \ie the barrier between $\theta_A, \theta_B \in \Theta$ is $\barrier{\theta_A}{\theta_B} := \max_{t \in [0, 1]} \widetilde{\mathcal{L}}_t(\theta_A, \theta_B)$, where

\begin{align}
\widetilde{\mathcal{L}}_t(\theta_A, \theta_B) :=& \,\genericloss{(1-t)\cdot\theta_A + t\cdot\theta_B} - \left((1-t)\cdot\genericloss{\theta_A} + t\cdot\genericloss{\theta_B} \right)
\vspace{-1em}
\label{eq:loss_barrier_definition}
\end{align}

is the difference between the loss value at $t$, and the linear interpolation of the losses at the end-points.
\setlength{\columnsep}{1em}
\setlength{\intextsep}{0em}
\begin{wrapfigure}{r}{0.25\linewidth}
\vspace{0em}
\includegraphics[width=\linewidth]{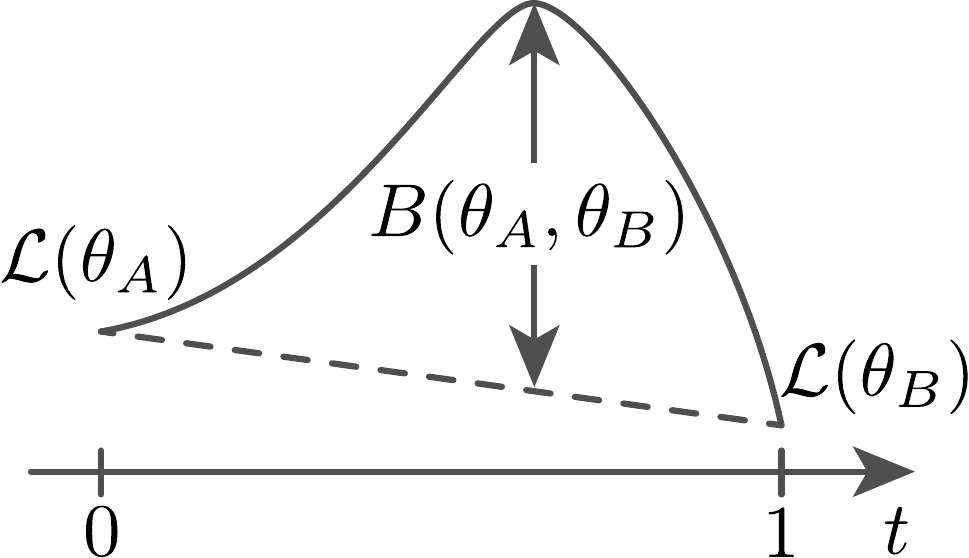}
\label{figure:loss_barrier}
\vspace{-1.2em}
\end{wrapfigure}
Two solutions $\theta_A,\theta_B\in\Theta$ are said to be \textbf{linearly mode-connected}, or \textbf{LMC} \citep{frankle2020lottery}, when their loss barrier is 
approximately zero: $\barrier{\theta_B}{\theta_A} \approx 0$.

The convexity conjecture is constructed upon a parameter space where the permutation symmetries are factored out. 
A \textbf{permutation invariance} \citep{brea2019weight} can be formulated as an equivalence relation $\sim$ between two points $\theta_A, \theta_B$ in the parameter space such that $\theta_A\sim\theta_B$ if and only if there exists a permutation $\pi$ of the parameters such that $\pi(\theta_A) = \theta_B$ \emph{and} the functions represented by them are identical: $f_{\theta_A}(x)= f_{\theta_B}(x)$ for all $x$. Given two points $\theta_A$ and $\theta_B$, we look for the permutation of $\theta_B$ that connects it to $\theta_A$ (or vice versa) with as low a loss barrier as possible \citep{ainsworth_git_re_basin_2022, entezari_permutation_invariances_2022, sinkhorn_re_basin_2023}. A \textbf{winning permutation}  \citep{entezari_permutation_invariances_2022} for models $\theta_A$ and $\theta_B$ is defined as 

\begin{align}
    \winperm{\theta_A}{\theta_B} 
    := 
    \argmin_{\pi \in \allperms{\theta_A}}
    \barrier{\pi(\theta_A)}{\theta_B}
\end{align}
where $\allperms{\theta}:=\{\pi\,|\,\pi(\theta)\sim\theta\}$ is the set of all function-preserving permutations of $\theta$.

\begin{conjecture}
\label{conjecture:convexity}
\textbf{Convexity Conjecture} \citep{entezari_permutation_invariances_2022}.
    Let $\solutionset{}$ be the set of SGD-reachable solutions for a deep neural network $f(\theta)$ trained for a certain task.
    Let $\theta_A,\theta_B\in\solutionset{}$ be two solutions. Then, there exists a minimum width $h$ such that if $f(\theta)$ is wider than $h$, then with high probability, $\theta_B$ can be permuted to obtain $\tilde{\theta}_B = \winperm{\theta_B}{\theta_A}(\theta_B)$ such that $\theta_A$ and $\tilde{\theta}_B$ are highly likely to be linearly mode-connected, \ie $\barrier{\tilde{\theta}_B}{\theta_A} \approx 0$.
\end{conjecture}
We refer to this as the (quasi-) convexity \cite{ainsworth_git_re_basin_2022, xiao2023compact} conjecture, because, by definition, a convex set is precisely a set where the line segment between any two elements is included in the set. The conjecture provides a geometric intuition that the solution set is generally convex, modulo permutations.

Theoretical results only validate the conjecture in limited settings, given sufficiently wide networks \citep{entezari_permutation_invariances_2022, ferbach2024proving}. Empirical validations exhibit mixed accounts. \cite{ainsworth_git_re_basin_2022} notably achieve zero barrier between two ResNet-20-32 models trained on CIFAR-10, but there remains a loss barrier between narrower models, even after weight matching. They further report network depth and dataset complexity as aggravating factors. \cite{benzing_random_2022} provide interesting insights using their activation-matching permutation algorithm. While fully connected networks (FCNs) live in the same loss valley even at initialization, convolutional nets (CNNs) are usually not connected even after considering permutation invariances. \cite{sinkhorn_re_basin_2023} introduce Sinkhorn re-basin, a differentiable permutation-finding approach; however, even with two-layer NNs, the barrier between CIFAR-10 models, albeit low, remains non-zero. For CNN architectures like VGG, the barrier is substantially high. \cite{altintas_disentangling_2023} show that aggravating factors for LMC include the Adam optimizer \citep{kingma2014adam}, absence of warmup, and task complexity.

Hence, in cases where strong evidence for the convexity conjecture is absent, it is important to consider other possible topologies for general DNN solution sets. 
To this end, we propose the star domain conjecture for characterizing DNN solution sets that do not enjoy convexity \cite{entezari_permutation_invariances_2022} modulo permutations.

\subsection{The star domain conjecture}
\label{section:star_domain_conjecture}

We propose a weaker form of convexity for characterizing DNN solution sets.
We argue that DNN solution sets are generally \textit{star domains}, modulo function-preserving permutations. While \cite{annesi2023star} demonstrates this property for simple spherical negative perceptrons (without permutations), we argue that it holds for even deeper, more complex nets after considering permutation invariances.

We start with the necessary definitions to make a formal description of the conjecture. A set $A \subset \mathbb{R}^n$ is a \textbf{star domain} if there exists an element $a_0 \in A$ such that for any other element $a \in A$ and $0\leq t\leq 1$, $(1-t)\cdot a_0+t\cdot a\in A$, \ie all points on the line segment between $a_0$ and $a$ lie in $A$. We call such $a_0$ a \textbf{star point}. In the context of the parameter space, we refer to the star point of a star-domain-shaped solution set as a \textbf{star model}.

\begin{conjecture} 
\label{conjecture:star_domain}
\textbf{Star Domain Conjecture}.
Consider the set $\solutionset{}$ of SGD-reachable solutions to a deep neural network $f(\cdot)$ trained to execute a certain machine learning task. Let $h$ be the minimum width for which $\solutionset{}$ becomes convex modulo permutations. Then there exists a constant $0<\alpha<1$ such that if $f(\cdot)$ is wider than $\alpha h$, then $\solutionset{}$ is highly likely to be a star domain modulo permutation symmetries, \ie there exists a \emph{star model} $\starmodel{} \in \solutionset{}$ such that for any other solution $\theta \in \solutionset{}$, it is possible to obtain $\tilde{\theta} = \winperm{\theta}{\starmodel{}}(\theta)$ such that $\barrier{\tilde{\theta}}{\starmodel{}} \approx 0$.

\end{conjecture}

A convex set is a special case of a star domain, where all the elements are star points. The star domain conjecture is thus a relaxation of the convexity conjecture.

\subsection{Finding a star model}
\label{section:ourmethod}

We provide empirical evidence for the star domain conjecture via two steps. First, we present a method for finding a star model. Second, we verify that the model found is indeed a star model: it has a low loss barrier with an arbitrary solution in $\solutionset{}$. Here, we focus on the first step.

We consider a necessary condition for a star model $\theta^\star$: given an arbitrary set of models $\sourcemodels = \{\theta_1, \theta_2, \dots, \theta_N\} \subset \solutionset{}$, $\theta^\star$ has to be connected to all of them, modulo permutation invariances. 

We present a recipe for finding such a $\starmodel{}$.

We first obtain a finite set $\sourcemodels = \{\theta_1, \theta_2, \dots, \theta_N\}$ of models, independently trained with different random seeds controlling the initialization, batch composition, and augmentation. We then formulate a loss function that, for fixed $\sourcemodels$, encourages low loss barriers between $\theta$ and some permuted versions of $\{\theta_1, \theta_2, \dots, \theta_N\}$. The objective may be expressed as
\begin{align}
    \starmodel{\sourcemodels} = \argmin_{\theta}\frac{1}{N}\sum\limits_{\theta_n \in \sourcemodels}\barrier{\theta}{\winperm{\theta_n}{\theta}(\theta_n)}
\end{align}
where $\winperm{\theta_n}{\theta}$ is the winning permutation defined in Section~\ref{section:background_convexity} that permutes $\theta_n$ without changing the represented function while minimizing the loss barrier against $\theta$. To solve this optimization problem, we propose to minimize the expected loss on the linear interpolation between the model in question $\theta$ and each source model $\theta_n$, after permutations. We modify the training objective as $ \starmodel{\sourcemodels} = \argmin_{\theta} \sdloss{\theta}{\sourcemodels}$
where

\vspace{-1.5em}
\begin{align}
    \sdloss{\theta}{\sourcemodels}  
    := \frac{1}{N}
    \sum_{n=1}^{N} 
    \int_{0}^{1} 
    \genericloss{(1-t)\cdot\theta + t\cdot\winperm{\theta_n}{\theta}(\theta_n)} 
    dt
\label{eq:star_model_main}
\end{align}

This expresses the expected loss on the set of line segments between $\theta$ and $\winperm{\theta_n}{\theta}(\theta_n)$, where each source model $\theta_n\sim\text{Unif}(\sourcemodels)$ is chosen at random and then each point on the line segment is sampled as $t \in \text{Unif}[0, 1]$. The optimization problem in \cref{eq:star_model_main} 
involves computational challenges.
Resolving the continuous integral over $t$ is non-trivial for complex learning problems.
Furthermore, $\winperm{\theta_n}{\theta}$ assumes access to the winning permutation. However, the winning permutation depends on $\theta$, which constantly changes during optimization. 
We introduce the following solutions.

\begin{wrapfigure}{R}{0.5\textwidth}
\begin{minipage}{0.5\textwidth}
\begin{algorithm}[H]
\caption{\ourmethod: Training a Star Model.}
\label{alg:star_training}
\KwIn{
dataset $\mathcal{D} = \{(x_i, y_i)\}_{i=1}^I$, 
source models $\sourcemodels = \{\theta_1, \theta_2, \ldots, \theta_N\}$,
initial model $\theta_0$, 
learning rate $\lambda$, 
number of batches $m$, 
number of steps $K$\;
Set $\theta \leftarrow \theta_0$.
}
\KwOut{$\theta$}
\For{$k=1$ \KwTo $K$}{
\If{$(k-1) \mod m == 0$}{
    \For{$n=1$ \KwTo $N$}{
        \textbf{Step 1.} Update $\theta_n \leftarrow \winperm{\theta_n}{\theta}(\theta_n)$\;
    }
}
{\noindent
\textbf{Step 2.} Sample $\theta_n \sim \text{Unif}(\sourcemodels)$, $t \sim \text{Unif}[0, 1]$, and a batch $\mathcal{B}$ from $\dataset$.\;\\
\textbf{Step 3.} Compute loss $\mathcal{L}((1-t)\cdot\theta + t\cdot\theta_n\,;\,\mathcal{B})$.\;\\
\textbf{Step 4.} Compute gradients $v \leftarrow \nabla_{\theta}\mathcal{L}((1-t)\cdot\theta + t\cdot\theta_n)$.\;\\
\textbf{Step 5.} Update $\theta \leftarrow \theta - \lambda(1-t) \cdot v$.\;}
}

\end{algorithm}
\vspace{-1em}
\end{minipage}
\end{wrapfigure}

\paragraph{Monte-Carlo optimization scheme.} 
Instead of estimating $\sdloss{\theta}{\sourcemodels} $ precisely at every iteration, we rely on a Monte-Carlo estimation scheme, inspired by the parameter-curve fitting method by \cite{garipov_mode_connectivity_2018}. At iteration $k\geq 1$, 
we sample $\theta_{n^{(k)}}$ uniformly from $Z$
and $t^{(k)}$ from $\text{Unif}[0, 1]$ (see \cref{appendix:different_sampling_schemes} for ablations using alternative sampling schemes). Hence, we obtain a single point on the manifold, calculate the cross-entropy loss at this point, and subsequently the gradients for updating $\theta$.

\paragraph{Finding optimal permutations.} We perform weight matching \citep{ainsworth_git_re_basin_2022}, \ie we seek a permutation $\pi_n$ that maximizes the dot product $\theta{} \cdot \pi_n(\theta_n)$, for each $\theta_n \in \sourcemodels$. This procedure aligns each source model $\theta_n$ with the candidate star model $\theta$. This operation is performed at the beginning of every \textit{epoch} instead of every \emph{iteration}, significantly speeding up the optimization process.

\cref{alg:star_training} describes the detailed procedure. Once we find a $\theta$ that has a low expected loss $\sdloss{\theta}{Z}$ on the linear paths to a finite set of source models $Z$, we may verify if this $\theta$ is likewise linearly connected with an arbitrary solution $\theta_{N+1}\notin Z$.  

\subsection{Empirical evidence} 
\label{section:star_domain_empirical}

We introduced \ourmethod to find a candidate star model. Now, we propose a method to verify if the model found in Section \ref{section:ourmethod} is a star model by checking its linear connection to an arbitrary solution $\theta_{N+1}\notin Z$, \ie not part of the set of source models used for finding the star model. We refer to such models as \textit{held-out} solutions $\heldout$ that are disjoint from the source models: $\heldout \cap \sourcemodels = \emptyset$. 

\begin{table}

\renewcommand{\arraystretch}{1.2}
\setlength{\tabcolsep}{.4em}
\captionsetup{font=small}
\caption{
\textbf{Empirically verifying the star domain conjecture.} 
\quotes{Regular loss} and \quotes{Star loss} indicate training losses for regular models in $\sourcemodels{}$ and star models $\starmodel{}$, respectively. 
\quotes{Star-regular} refers to the barrier $\barrier{\starmodel{}}{\theta_h}$ between a star model and one of the heldout models in $\heldout$. 
For comparison, \quotes{Regular-regular} is the loss barrier $\barrier{\theta_A}{\theta_B}$ between two arbitrary models.  
We report values up to one standard deviation over several runs, except for ImageNet. 
In each case, star models exhibit significantly lower loss barriers with other models, than the corresponding average loss barrier between two regular models.
}
\label{table:star_model_empirical_verification}
\vspace{.5em}
\centering
\small
\begin{tabular}{lccccccr}
\toprule
\normalfont{Dataset} &
\normalfont{Architecture} &
\normalfont{Regular loss} &
\normalfont{Star loss} &
\normalfont{Regular-regular} &
\normalfont{Star-regular} &
\\
\midrule
CIFAR-10 &
ResNet-18 &
$0.001 \pm 0.000$ &
$0.001 \pm 0.000$ &
$0.383 \pm 0.056$ &
$0.078 \pm 0.007$ &
\\
CIFAR-10 &
ResNet-18 (Adam) &
$0.001 \pm 0.000$ &
$0.015 \pm 0.000$ &
$1.368 \pm 0.551$ &
$0.335 \pm 0.022$ &
\\
CIFAR-10 &
VGG11 &
$0.003 \pm 0.000$ &
$0.022 \pm 0.000$ &
$0.515 \pm 0.034$ &
$0.131 \pm 0.005$ &
\\
CIFAR-10 &
VGG19 &
$0.001 \pm 0.000$ &
$0.059 \pm 0.000$ &
$1.281 \pm 0.153$ &
$0.336 \pm 0.078$ &
\\
CIFAR-10 &
DenseNet &
$0.001 \pm 0.000$ &
$0.157 \pm 0.000$ &
$4.634 \pm 0.727$ &
$1.729 \pm 0.409$ &
\\
\midrule
CIFAR-100 &
ResNet-18 &
$0.004 \pm 0.001$ &
$0.005 \pm 0.000$ &
$2.905 \pm 0.047$ &
$0.756 \pm 0.049$ &
\\
CIFAR-100 &
DenseNet &
$0.006 \pm 0.000$ &
$0.635 \pm 0.000$ &
$6.920 \pm 0.216$ &
$3.735 \pm 0.180$ &
\\
\midrule
ImageNet-1k &
ResNet-18 &
$0.711$ &
$1.380$ &
$5.948$ &
$2.794$ &
\\
\bottomrule
\end{tabular}
\end{table}

\begin{figure*}[t]
    \centering
    \small
    \setlength{\tabcolsep}{.1em}
    \begin{tabular}{cc}
        \includegraphics[width=.49\linewidth]{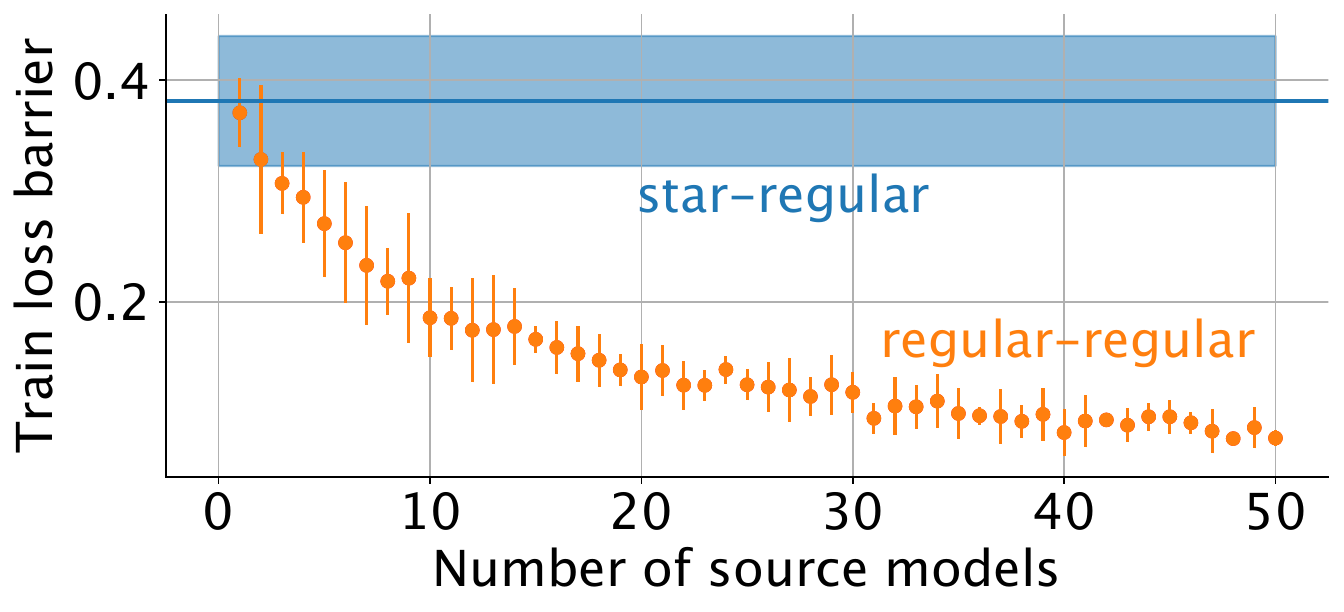} &
        \includegraphics[width=.49\linewidth]{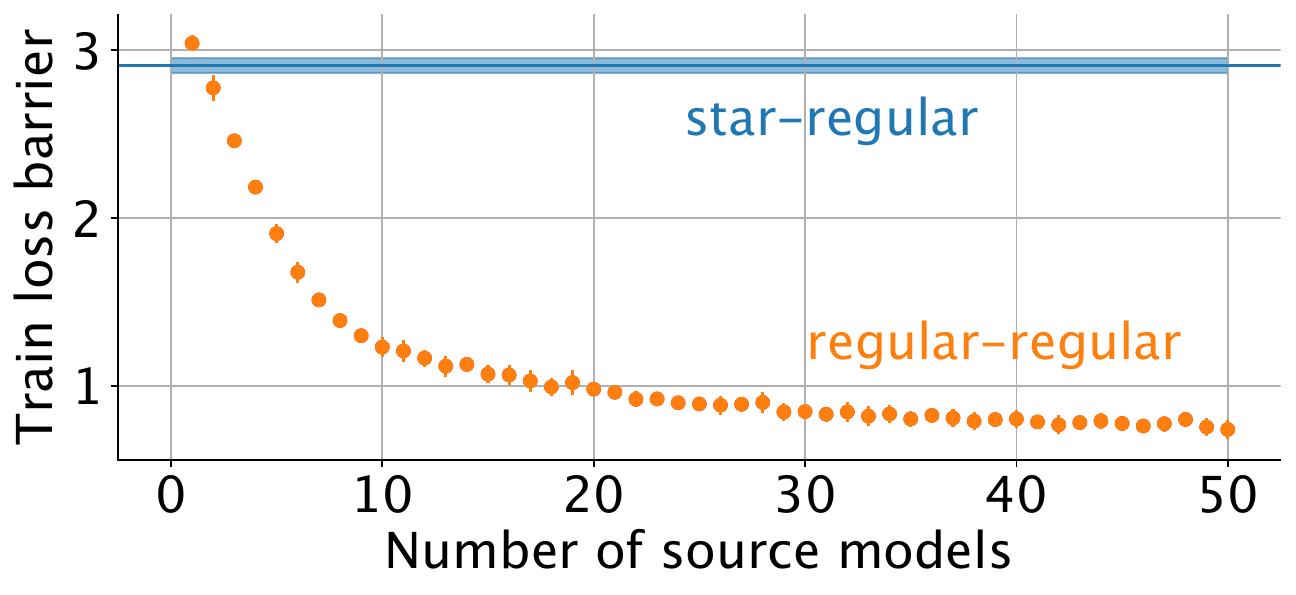} 
        \\
    \end{tabular}
    \caption{\textbf{Starness of a star model vs source models.} We plot the loss barriers $B(\theta^\star,\theta_h)$ between star models $\theta^\star$ and heldout models $\theta_h\in H$ at different numbers of source models $Z$ used for learning the star model $\theta^\star$ (\starcolor{orange points}). The heldout set is disjoint with the source models: $H\cap Z=\emptyset$. We provide a reference point given by the loss barrier between two regular solutions $B(\theta_A,\theta_B)$ for $\theta_A,\theta_B\in\solutionset{}$ (\regcolor{blue plot}). The error bars indicate one standard deviation across five held-out models $|H|=5$.
    Incorporating more source models $|Z|$ enables finding a better star model with a lower loss barrier against an arbitrary solution.}
    \vspace{-2em}
    \label{figure:cifar_resnet_loss_barrier_vs_num_anchors}
\end{figure*}

\begin{figure*}[t]
    \centering
    \small
    \setlength{\tabcolsep}{.1em}
    \begin{tabular}{lc}
        \includegraphics[width=.49\linewidth]{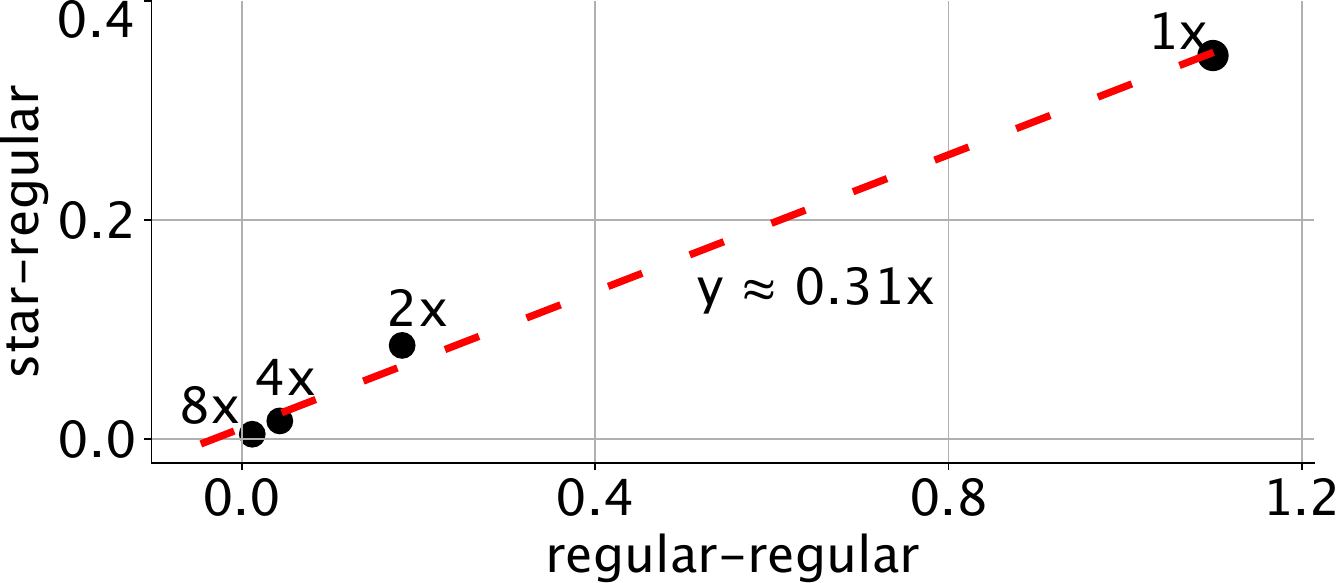} &
        \includegraphics[width=.49\linewidth]{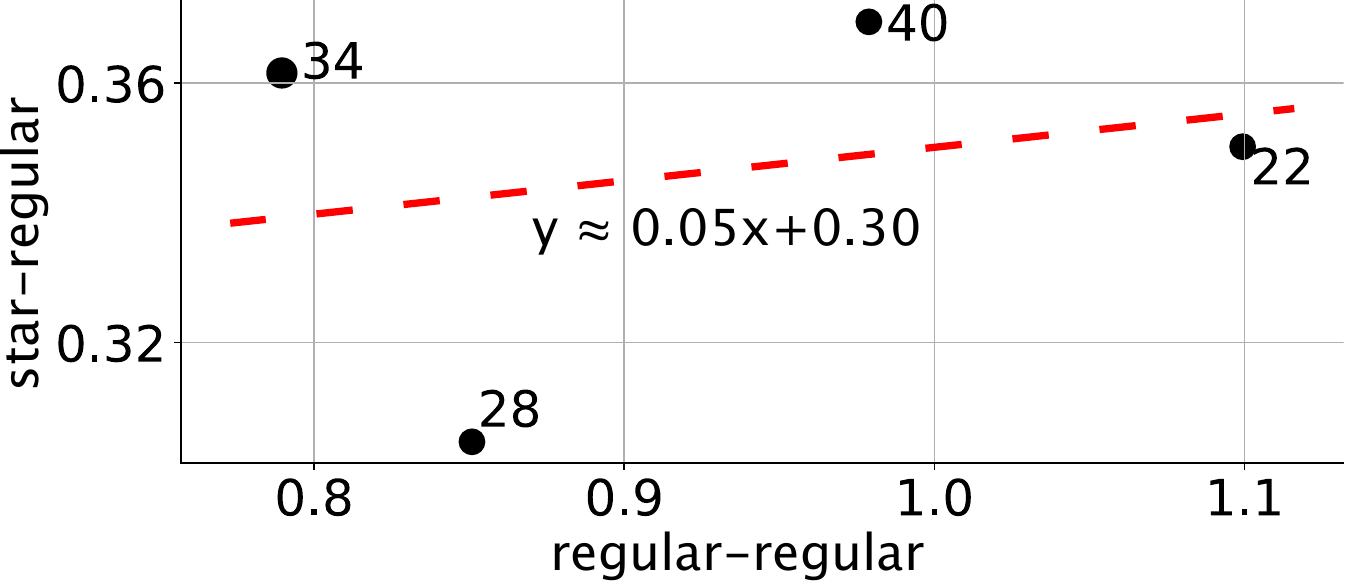} 
        \\
         \multicolumn{1}{c}{Width} &
         \multicolumn{1}{c}{Depth}
        \\
    \end{tabular}
    \caption{
    \textbf{\quotes{Starness} vs. model width and depth}. For starness vs. model width (left), we vary the width of a WideResNet (depth $22$) from $1\times$ to $8\times$. For starness vs. model depth, we vary the depth of a WideResNet (width $1\times$) from $22$ to $40$ layers. For each depth-width combination, we plot the loss barriers $B(\theta^\star,\theta_h)$ between star models $\theta^\star$ and heldout models $\theta_h\in H$ on the y-axis. As a reference point, we plot the barrier between two regular solutions $B(\theta_A,\theta_B)$, on the x-axis. The points are annotated with the corresponding widths or depths. Star models consistently enjoy better linear connections with regular models, than do the regular models amongst each other. 
    }
    \label{figure:cifar_wrn_width_depth}
\end{figure*}

\begin{figure*}[t]
    \centering
    \small
    \setlength{\tabcolsep}{.1em}
    \begin{tabular}{ccccc}
        \rotatebox[origin=b]{90}{\small CIFAR10\hspace{-5em}} & 
        \includegraphics[width=.24\linewidth]{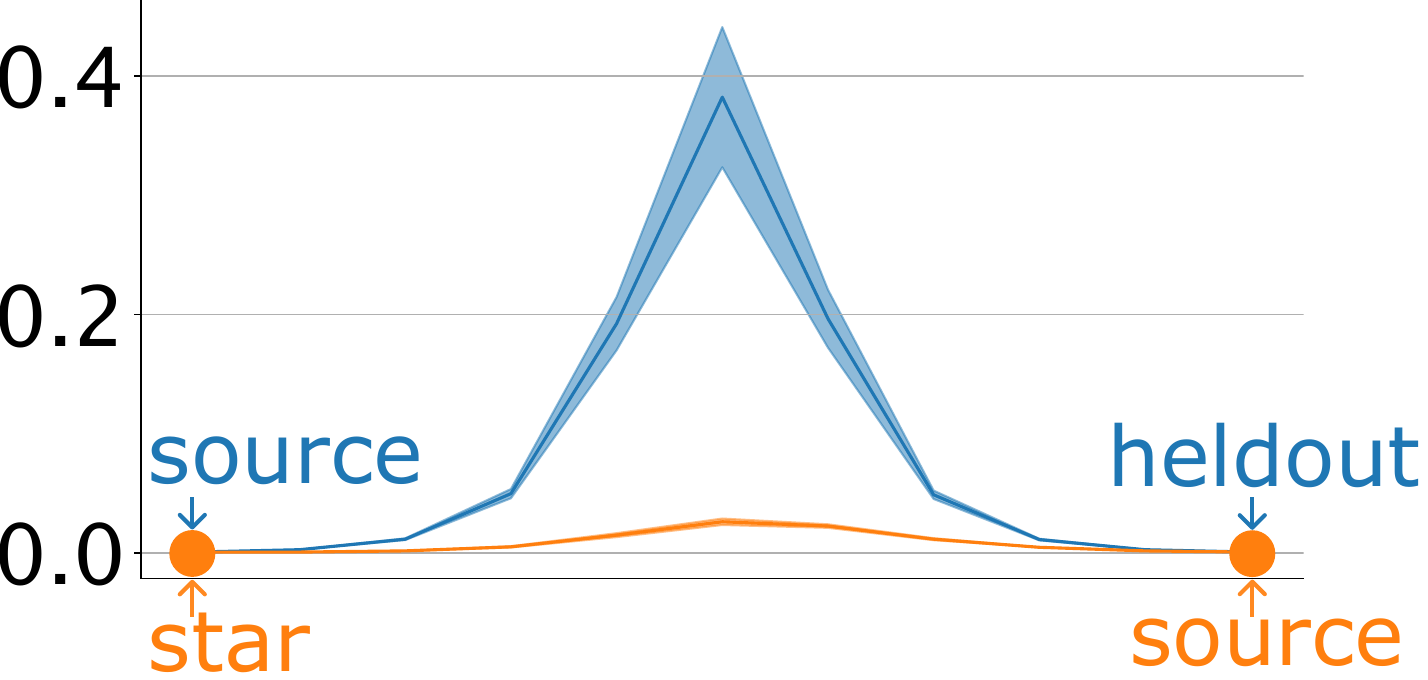} &
        \includegraphics[width=.24\linewidth]{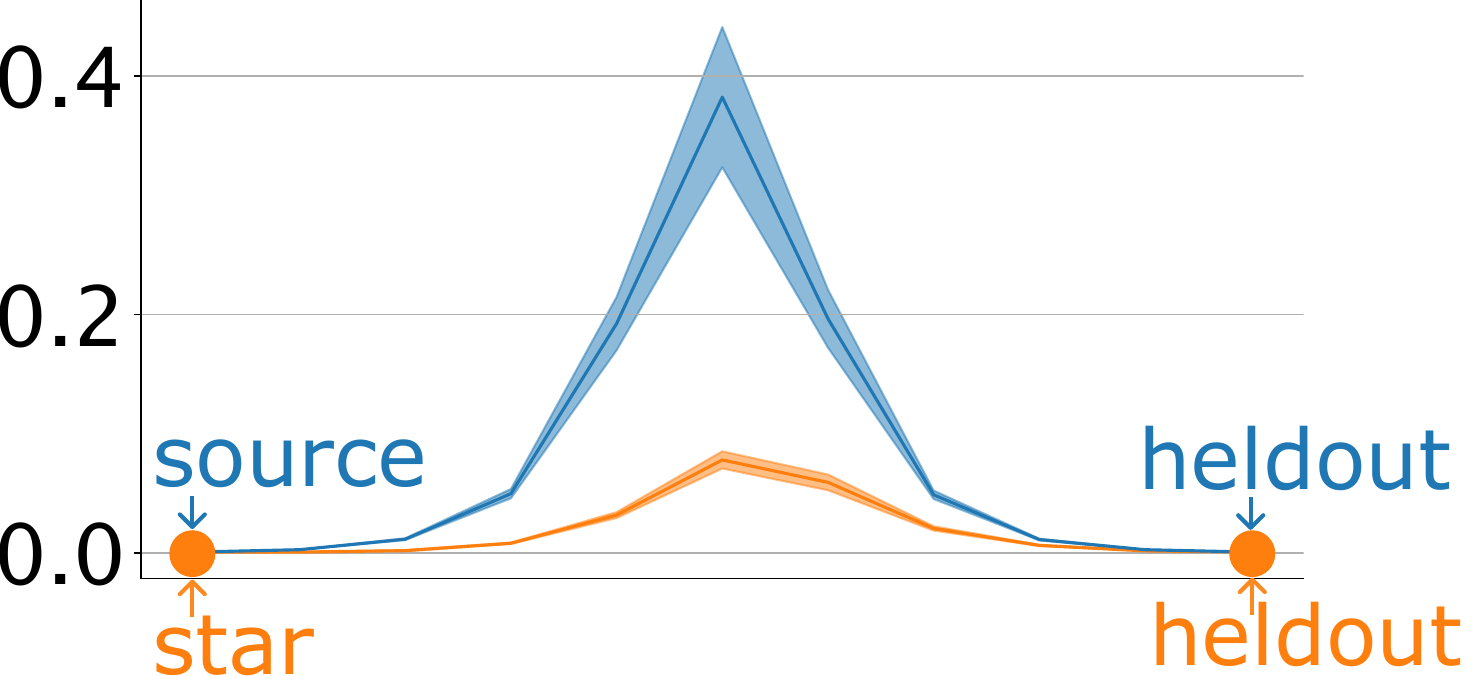} &
        \hspace{0.3em}
        \includegraphics[width=.24\linewidth]{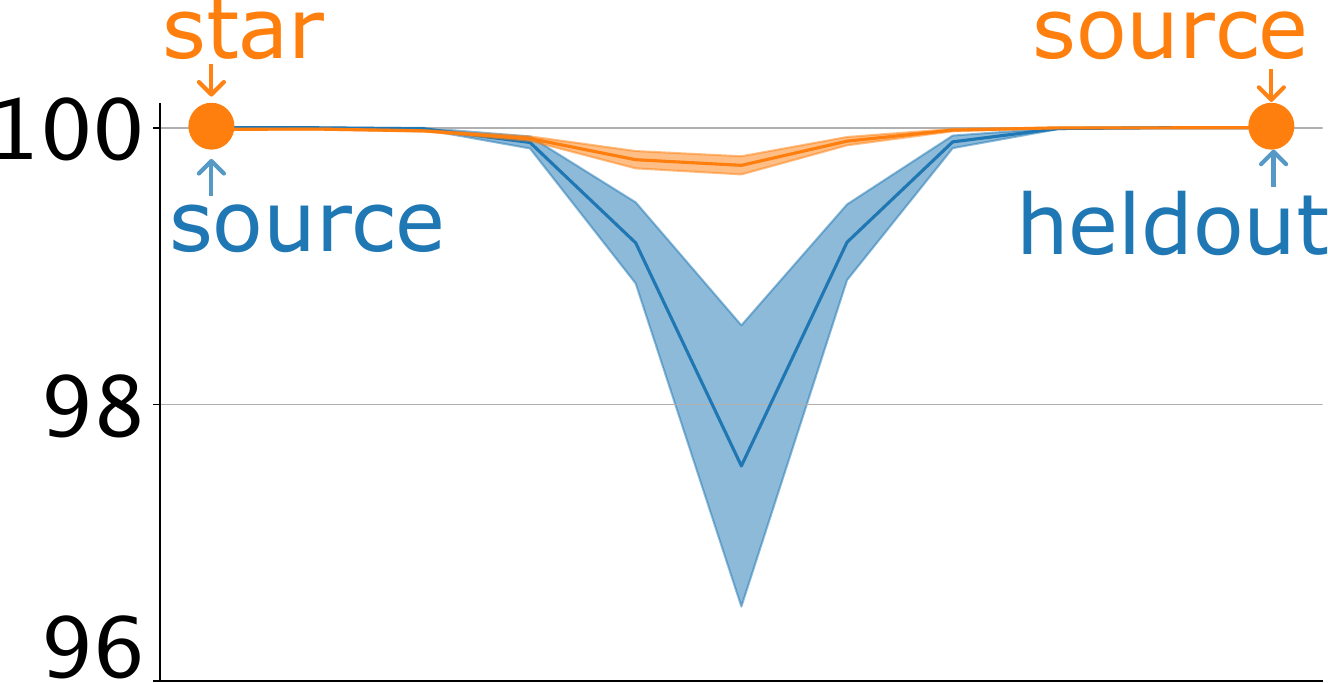} &
        \includegraphics[width=.24\linewidth]{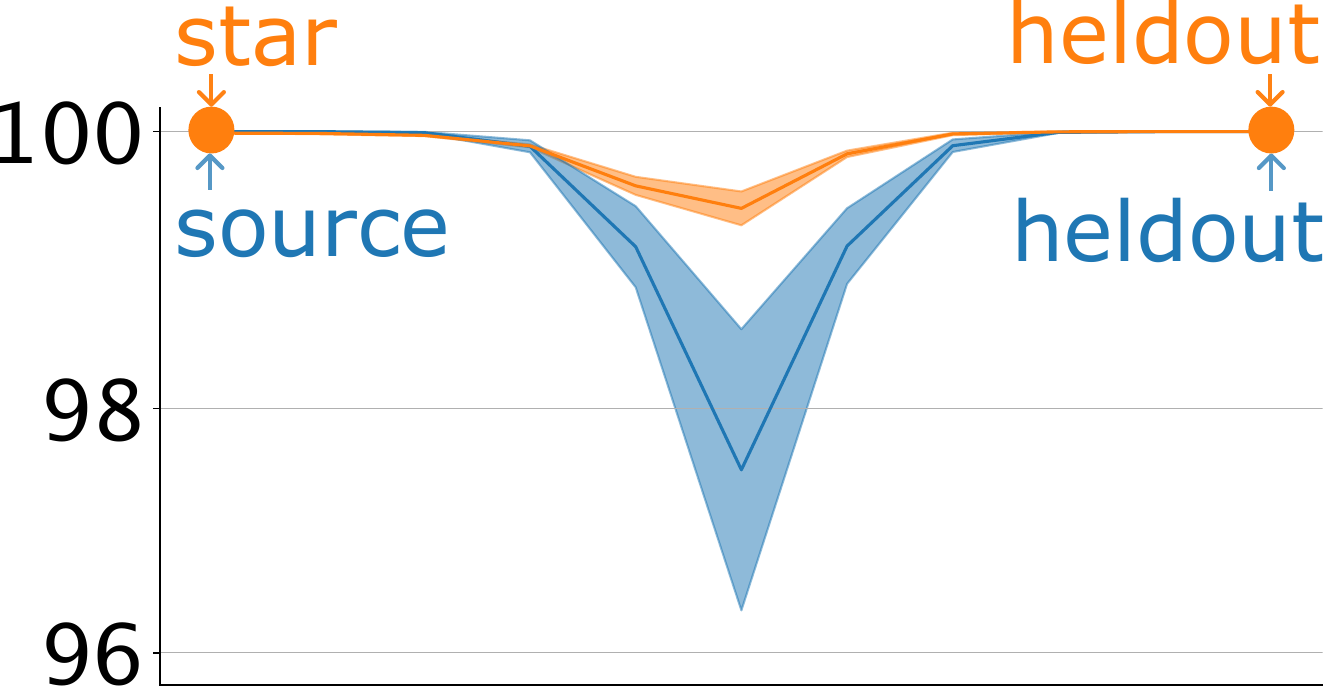}
        \\
        \rotatebox[origin=c]{90}{\small CIFAR100\hspace{-5em}} & 
        \includegraphics[width=.24\linewidth]{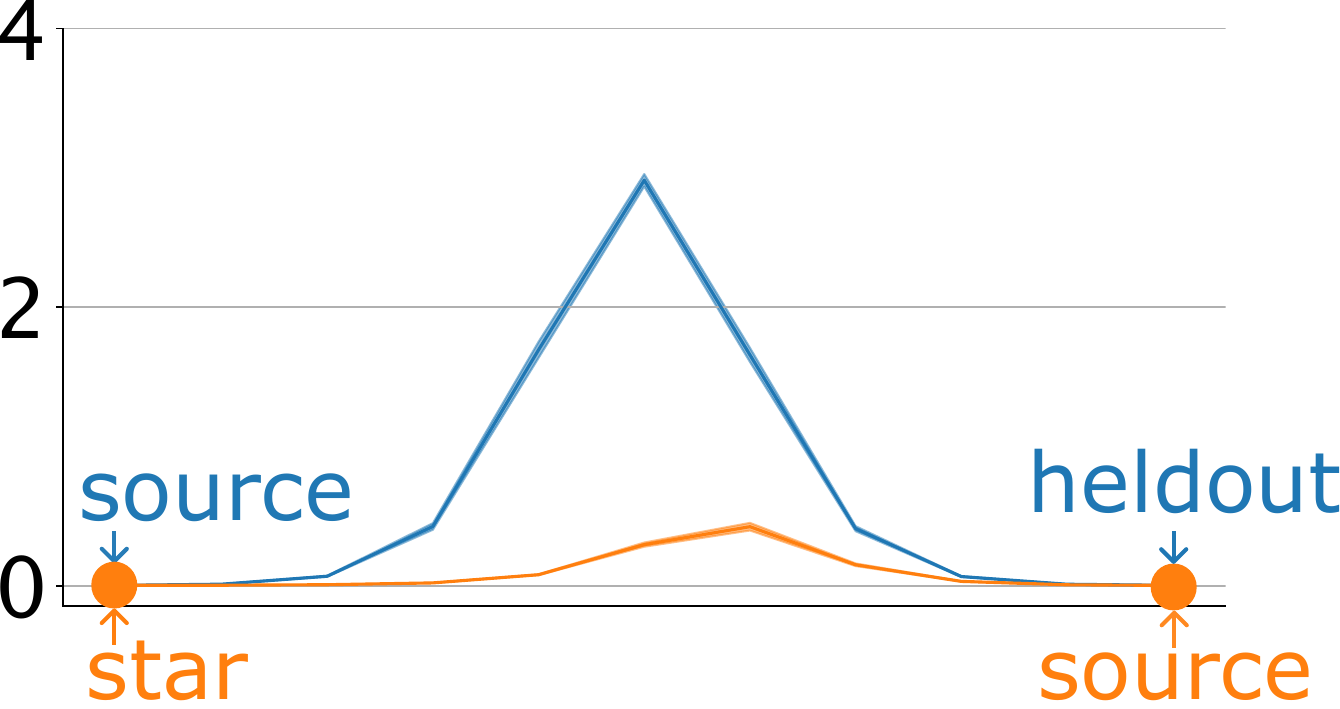} &
        \includegraphics[width=.24\linewidth]{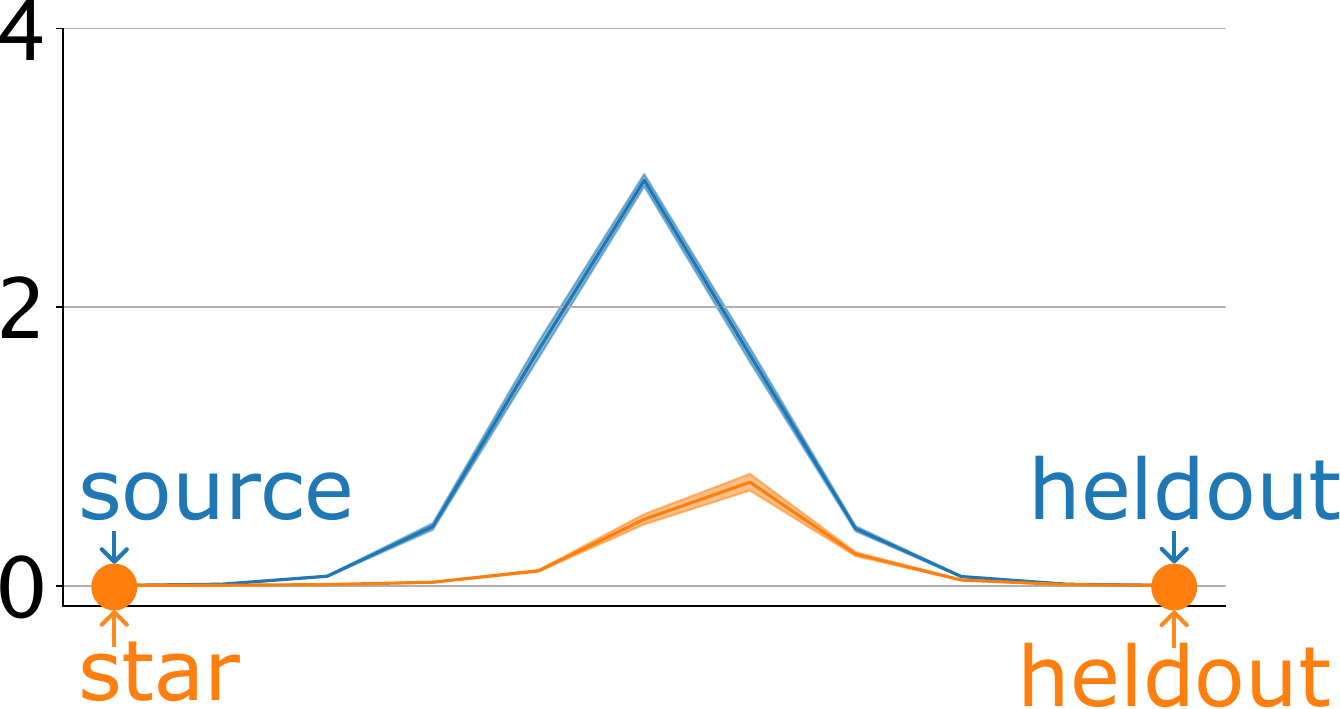} &
        \hspace{0.3em}
        \includegraphics[width=.24\linewidth]{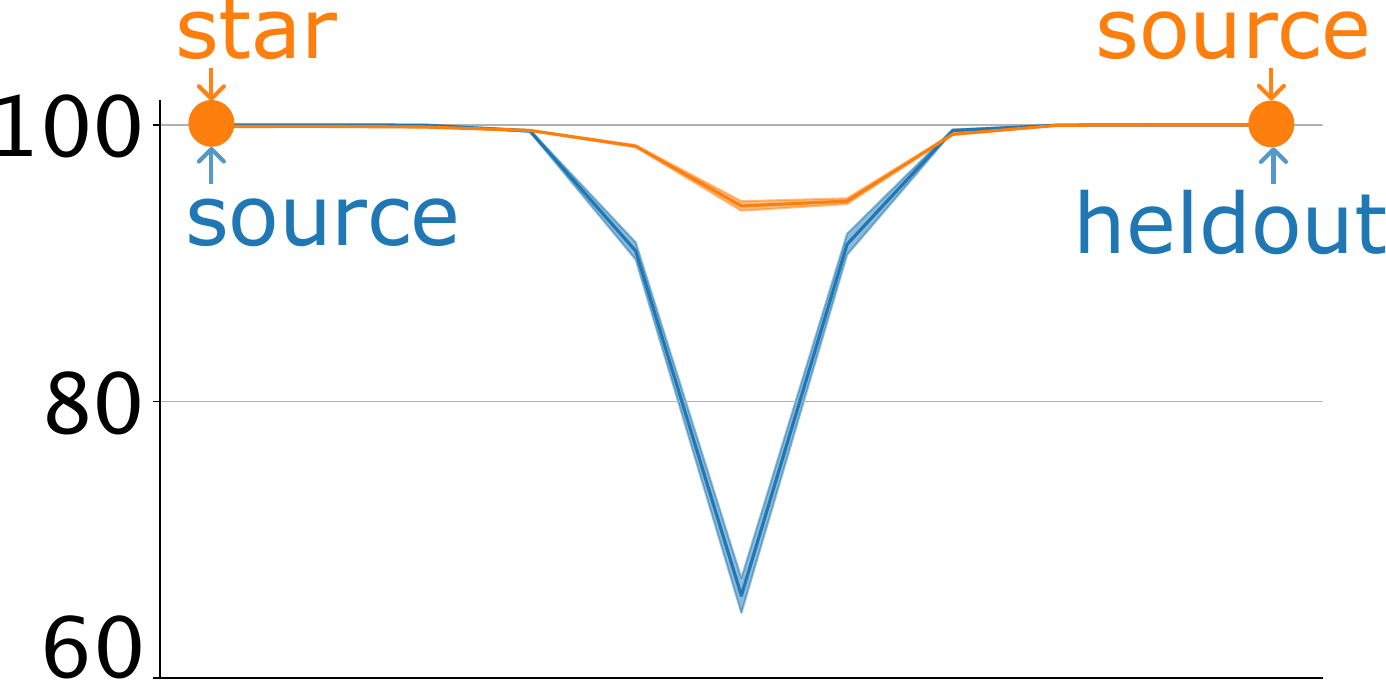} &
        \includegraphics[width=.24\linewidth]{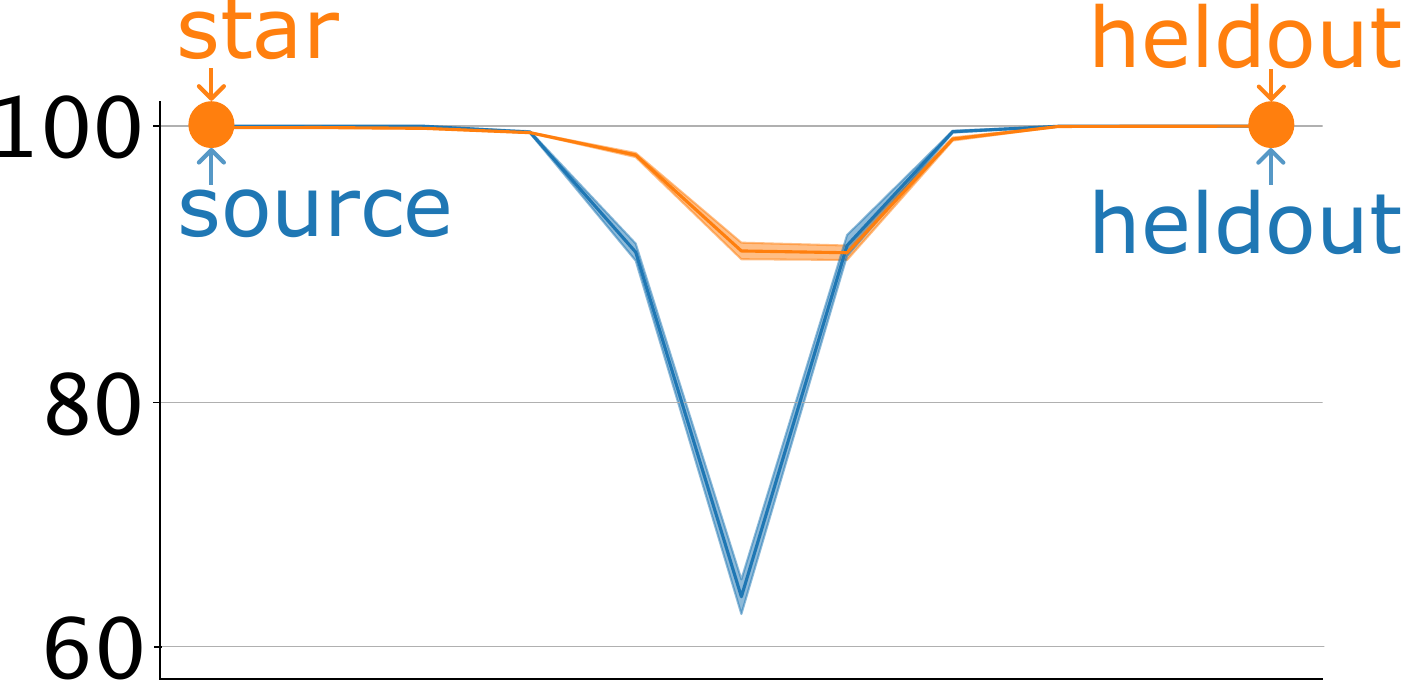}
        \\
        & \multicolumn{2}{c}{Train loss}
        & \multicolumn{2}{c}{Train accuracy}
    \end{tabular}
    \caption{\textbf{Loss barriers for star models}. We interpolate between a star model $\theta^\star$ and regular models that are trained with SGD. There are two types of regular models, depending on whether they are used for finding the star model: source models $Z$ are used, and heldout models $H$ are not. Along the interpolation, we visualize the loss barrier by plotting the loss and accuracy values (\starcolor{orange curves}). For these curves, $t=0$ corresponds to the star model $\theta^\star$. For reference, we plot the interpolation between two arbitrary regular models (\regcolor{blue curves}). The error bands correspond to one standard deviation.}
    \vspace{-15pt}
    \label{figure:cifar_resnet_interpolation_plots}
\end{figure*}

We describe our main findings with reference to ResNet-18 \citep{resnet_he_2016} models trained on CIFAR \citep{krizhevsky2009learning} using SGD, using $50$ source models and $5$ held-out models. We present results for additional architectures (\eg VGG \citep{simonyan_very_2015} and DenseNet \citep{huang_densely_2018}), a large-scale dataset (ImageNet-1k \cite{deng2009imagenet}) and settings (for instance, Adam \cite{kingma2014adam}) in \cref{table:star_model_empirical_verification} and \cref{appendix:additional_results}. 
Likewise, our empirical findings are built upon the training loss and accuracy, but we confirm that they also transfer over to test loss and accuracy in \cref{appendix:test_metrics}. We largely use standard recipes to train the models in our experiments, with the exception of star models where we additionally incorporate the steps in \cref{alg:star_training}. We further describe our experimental setup in \cref{appendix:implementation_details}. We summarize our observations below. 

\textbf{Convexity conjecture does not hold.}
In \cref{figure:cifar_resnet_interpolation_plots}, we show loss barriers between two independently trained solutions (blue ``regular-regular'' curves). 
We observe that the loss increases and accuracy drops significantly at around $t=0.5$, even after applying the algorithm \cite{ainsworth_git_re_basin_2022} to find the winning permutation.
We present another piece of evidence that the convexity conjecture does not hold for thin ResNets, reconfirming the findings of \cite{ainsworth_git_re_basin_2022}.

\textbf{Star model has low loss barriers with other solutions.} 
In \cref{figure:cifar_resnet_interpolation_plots}, 
we show the training losses and accuracies along linear paths between the candidate star model $\theta^\star$ and other types of solutions (either source models $\sourcemodels$ or held-out models $\heldout$). They are indicated with red curves. As a reference, we always plot the confidence interval of loss and accuracy values along the line segments between two regular solutions (blue curves). 
For the source models in $Z$, star-to-regular connections enjoy essentially zero loss barriers, in contrast with regular-to-regular connections, which remain significantly higher at $0.381$, for CIFAR-10. 
This demonstrates that it is possible to find a model $\starmodel{}$ simultaneously connected to $|\sourcemodels|=50$ models. 
The same is true for the line segments between the star model $\starmodel{}$ and a held-out model picked from $|\heldout|=5$ models; even though the barrier between the star model and the heldout model is non-zero, it remains as low as $0.077$ compared to $0.381$ for the regular-to-regular case.

\textbf{A greater number of source models enhances \quotes{starness}.} 
Our star model is constructed from the set of source models $\sourcemodels$. 
We question whether greater $|\sourcemodels|$ induces greater ``starness'' of the solution found by \ourmethod. 
In \cref{figure:cifar_resnet_loss_barrier_vs_num_anchors}, we plot the loss barrier against the number of source models $2\leq |\sourcemodels|\leq 50$ used to construct the star model.
For statistical significance, we include loss barrier statistics between two regular, independently trained models in $\solutionset$ with error bars indicating one standard deviation. 
We observe that {the loss barriers between these star models and the held-out models decrease as $|\sourcemodels|$ increases.} The decreasing trend has not saturated after $|\sourcemodels|=50$. We stopped there because of computational limits. However, including more source models is likely to enhance connectivity between the obtained star model and the other solutions even further.

\textbf{Effect of model width and depth.}
Prior work stresses the importance of model width and depth \cite{ainsworth_git_re_basin_2022, entezari_permutation_invariances_2022} in determining loss barriers between two solutions. We investigate the effect of model width and depth for residual nets. Specifically, we consider WideResNets \cite{zagoruyko_wide_resnets_2017} of widths $1\times$, $2\times$, $4\times$, and $8\times$ that of a normal ResNet (depth $22$). We also consider ResNets of depths $22$, $28$, $34$, and $40$. We compare the barriers achieved by \quotes{regular-regular} and \quotes{star-regular} pairs for each case \cref{figure:cifar_wrn_width_depth}. Our investigation confirms existing reports of decreasing loss barriers as model width increases. We observe significantly lower star-regular barriers than regular-regular barriers for models of identical widths (\eg roughly $0.004$ compared to $0.012$ at width $8x$). In fact, it is possible to fit a linear regression line to the observed barrier values, wherein the star-regular barriers are about a third of the regular-regular barriers at any given width (\cref{figure:cifar_wrn_width_depth}, left). We draw similar conclusions from varying depth (\cref{figure:cifar_wrn_width_depth}, right), although the change in barriers as we change the model depth is not quite as pronounced as it is for the varying width case. This observation also motivates the statement of our conjecture, as we explain in \cref{appendix:barrier_relationships}.

\textbf{Effect of optimizer.}
While both the convexity conjecture and the star domain conjecture involve solution sets obtained through SGD, we also investigate the impact of using the Adam optimizer \citep{kingma2014adam}. Specifically, we train $15$ regular models, with $|\heldout|=5$ and ($|\sourcemodels|=10$). We then train a star model and evaluate its barriers with the models in the held-out set. Results can be found in \cref{table:star_model_empirical_verification} (second row). We observe that Adam-trained regular solutions have a higher loss barrier between them ($1.368$) compared to SGD-trained regular solutions ($0.383$). Likewise, the barrier between the star model and regular models also increases from $0.078$ for SGD solutions to $0.335$ for Adam solutions. While both \quotes{regular-regular} and \quotes{star-regular} connections suffer with this change of optimizer, \quotes{star-regular} connections still fare significantly better than \quotes{regular-regular} connections. This finding suggests that Adam solutions are also highly likely to enjoy star-shaped connectivity.

\textbf{Caveats}. Despite the promising observations above, our star domain conjecture is not theoretically validated and thus remains a conjecture. 
From the empirical perspective, loss barriers between the star model and other solutions often yield values that are significantly greater than zero. 
However, we emphasize that this paper is focused on providing the lower bound in evidence supporting the star domain conjecture.
Considering a larger number of source models for the star model construction, improving \ourmethod, and developing a better algorithm for finding the winning permutations will potentially contribute to the discovery of better star models in the solution set.

\textbf{Conclusion}.
Our experimental results confirm existing reports that the convexity conjecture requires very wide networks to hold, and has otherwise several failure cases for which we propose a relaxed version, \viz the star domain conjecture.
We obtain strong empirical evidence that the star model found through \ourmethod is likely to be a true star model.
Our analysis thus sheds further light on solution set geometry for narrower and deeper networks, as well as for complex learning tasks where the convexity conjecture struggles.
We invite the community to expand upon our findings and converge toward a more accurate understanding of the loss landscape.

\section{Practical applications}
\label{section:practical_applications}

The star domain conjecture introduces a novel dichotomy of solution types: \quotes{star} and \quotes{non-star} models. 
Most solutions are non-star and lack linear connections with other solutions. 
However, in \cref{section:star_domain_conjecture}, we have presented strong evidence for the existence of star models.
In this section, we examine the properties and potential benefits of star models in practice. 
\cref{subsection:bayesian_model_averaging} explores whether star models and the surrounding star domain provide a better posterior for Bayesian Model Averaging. In \cref{section:substitute_to_ensembling}, we propose star models as a practical alternative to model ensembling.

\begin{table}
\renewcommand{\arraystretch}{1.2}
\setlength{\tabcolsep}{.1em}
\small
\caption{\textbf{Model fusion performances}. 
\quotes{Regular} indicates single models; Ensemble indicates a vanilla average of the probability vectors across the member models; \quotes{Star} indicates a model found using \ourmethod using the regular models as the source set $\sourcemodels$. ResNet18 has been used throughout. We show one standard deviation for the error bars. In addition, we report the accuracy of the best member in the ensemble (\quotes{Best of $n$}) and the accuracy of the best star model (\quotes{Best of $3$}).
Star models perform better than single, regular models but use only a fraction of the compute required by the ensemble at test time.}
\label{table:cifar10_resnet18_model_fusion}
\vspace{.5em}
\centering
\begin{tabular}{lcccccccr}
\toprule
\multicolumn{2}{l}{\normalfont{Dataset \hspace{1.5em}\#\ignorespacesafterend Models}} &
\normalfont{Regular}&
\normalfont{Best of $n$}&
\normalfont{Ensemble} &
\normalfont{Star} &
\normalfont{Best of $3$} &
\\
\midrule

 &
$2$ &
$95.2 \pm 0.03$ &
$95.24$ &
$95.8$ &
$95.3 \pm 0.16$ &
$95.43$ &
\\
\normalfont{CIFAR-10} &
$5$ &
$95.1 \pm 0.14$ &
$95.24$ &
$96.0$ &
$95.2 \pm 0.15$ &
$95.27$ &
\\
&
$50$ &
$95.1 \pm 0.16$ &
$95.44$ &
$96.3$ &
$95.3 \pm 0.20$ &
$95.54$ &
\\
\midrule
 &
$2$ &
$77.3 \pm 0.16$ &
$77.49$ &
$79.6$ &
$78.0 \pm 0.24$ &
$78.14$ &
\\
\normalfont{CIFAR-100} &
$5$ &
$77.4 \pm 0.21$ &
$77.68$ &
$80.4$ &
$78.1 \pm 0.02$ &
$78.15$ &
\\
 &
$50$ &
$77.3 \pm 0.28$ &
$77.94$ &
$81.3$ &
$78.4 \pm 0.10$ &
$78.48$ &
\\  
\midrule
\multicolumn{2}{l}{\normalfont{Train / test complexity}} &
\multicolumn{2}{c}{$\mathcal{O}(1) / \mathcal{O}(1)$} &
$\mathcal{O}(n) / \mathcal{O}(n)$ &
\multicolumn{2}{c}{$\mathcal{O}(n) / \mathcal{O}(1)$} &
\\
\bottomrule
\end{tabular}
\end{table}

\begin{figure*}[t!]
    \centering
    \begin{subfigure}{0.32\textwidth}
        \includegraphics[width=\textwidth]{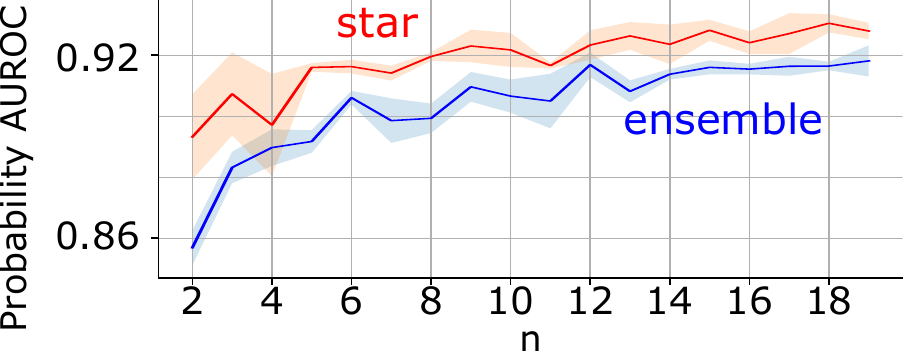}
        \caption{AUROC (Max. Probability)}
    \end{subfigure}
    \begin{subfigure}{0.32\textwidth}
        \includegraphics[width=\textwidth]{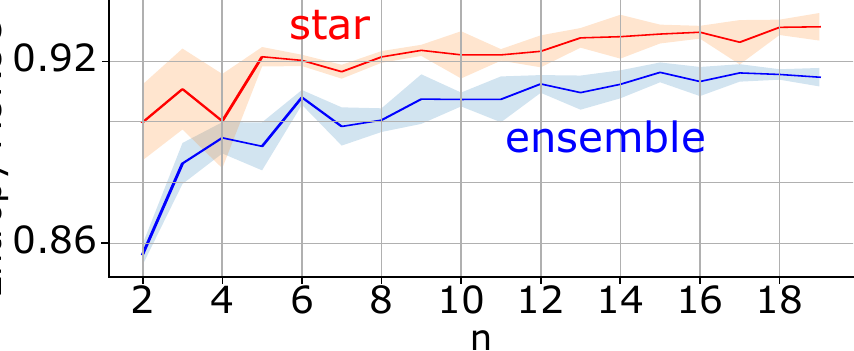}
        \caption{AUROC (Entropy)}
    \end{subfigure}
    \begin{subfigure}{0.32\textwidth}
        \includegraphics[width=\textwidth]{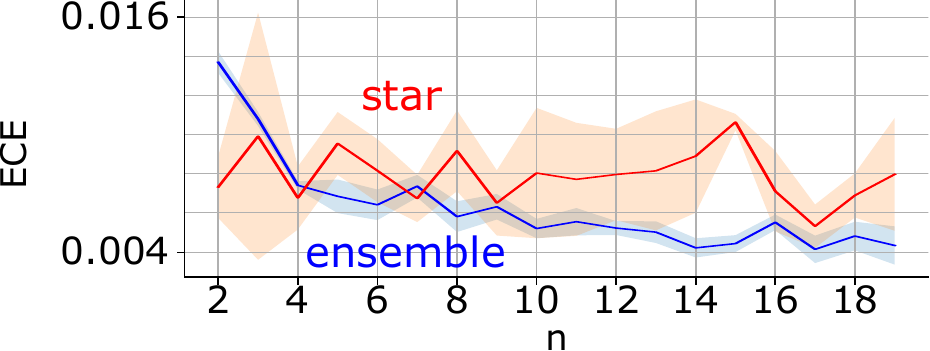}
        \caption{ECE}
    \end{subfigure}
    \caption{\textbf{Bayesian model averaging}. The star model was trained using 50 source models. The x-axis denotes the number of models sampled from the \textcolor{red}{star domain} for Bayesian model averaging or from the set of \textcolor{blue}{source models}.} 
    \vspace{-2em}
    \label{figure:bma_ciar10_resnet18}
\end{figure*}

\subsection{Bayesian model averaging}
\label{subsection:bayesian_model_averaging}

Bayesian model averaging (BMA) enhances uncertainty estimation by averaging predictions from the posterior of models in the parameter space. 
Posterior families in the literature range from simple Gaussian \citep{bbb} and Bernoulli \citep{dropout} distributions to more complex geometries like splines \citep{garipov_mode_connectivity_2018} and simplices \citep{benton_loss_simplexes_2021}. 
Here, we examine if the star domain provides a good posterior family for BMA-based uncertainty estimation.

\textbf{Setup}.
The posterior of interest is the collection of line segments between the star model $\starmodel{}$ and other solutions $\{\theta_1,\cdots,\theta_N\}$ that are independently found. 
Similarly to \ourmethod, we sample first from the model index of $\{1,\cdots, N\}$ uniformly and then sample from the line segment $\text{Unif}[0,1]$. 
As in standard BMA, we consider a set of models sampled from the posterior and the post-softmax average of these models.
We use ResNet-18 models trained on CIFAR-10. As a baseline, we present the BMA for the independent solutions $\{\theta_1,\cdots,\theta_N\}$.

\textbf{Evaluation}.
We assess the predictive uncertainty of the BMA-based confidence estimates. For the ranking metric, we use the area under ROC curve (AUROC). We consider both max-probability and entropy-based confidence measures. We also show results based on the expected calibration error (ECE).

\textbf{Results}.
\cref{figure:bma_ciar10_resnet18} shows uncertainty quantification at different numbers of posterior samples from $2$ to $19$. 
BMA using the star domain posterior consistently exhibits better AUROC values than baseline deep-ensemble estimates.
However, ECE is worse than that of the deep ensemble.
The star domain posterior provides avenues for more precisely ranked uncertainty estimates, albeit absolute-value uncertainty quantification may not be precise.

\textbf{Conclusion}.
Our proposed star domain posterior offers better uncertainty estimates than the deep ensemble baseline in rank-based predictive uncertainty evaluation.

\subsection{Potential usage in model fusion}
\label{section:substitute_to_ensembling}

Given a fixed amount of training data, a popular approach to maximize model generalizability is ensembling, \ie fusing predictions from multiple independent models. This basic approach suffers from computational complexities during both training and inference. Every input has to be processed by individual member models at test time. Storing multiple models also leads to a higher memory footprint, scaling linearly with the number of ensemble members. 

\ourmethod can also be understood as a method for aggregating multiple source models $Z=\{\theta_1,\cdots,\theta_N\}$ into a single model $\starmodel{}$. 
From a computational perspective, star models reduce the necessary time and storage complexity during inference.
We investigate whether the star models provide an enhanced generalization compared to the individual models. 
    
\textbf{Setup}.
We slightly modify the training objective of \ourmethod to align it with a better generalization capability of the star model. We add a cross-entropy term $\genericloss{\theta}$ so that $\mathcal{L}_{\text{total}}(\theta, \sourcemodels) = \sdloss{\theta}{\sourcemodels} + \genericloss{\theta}$, where $\sdloss{\theta}{\sourcemodels}$ is the original optimization objective for the star model discovery in \cref{eq:star_model_main}.

\textbf{Evaluation}.
We evaluate test accuracies for star models trained with varying numbers of source models ($|\sourcemodels|$) and compare them to ensembles using the same source models.

\textbf{Results}.
Results in \cref{table:cifar10_resnet18_model_fusion} show that star models consistently outperform regular models ($78.4\%$ vs. $77.3\%$) for CIFAR-100 with $|\sourcemodels|=50$). While less accurate than ensembles over $\sourcemodels$, star models require only a fraction of the compute during inference.

\textbf{Conclusion}.
\quotes{Starness} of a solution may enhance generalization. 
In scenarios where test-time inference costs are critical, star models could be a promising alternative to vanilla ensembles.

\section{Conclusion}
\label{section:conclusion}

This paper proposes a novel understanding of SGD loss landscapes. The traditional picture before \cite{garipov_mode_connectivity_2018} was one of extreme non-convexity, in contrast with the current picture of near-perfect convexity in a canonical, modulo-permutations space \citep{entezari_permutation_invariances_2022} for extremely wide nets. Our claim becomes relevant when narrower and deeper nets, complex datasets, and different optimization schemes are considered. We propose a weaker form of convexity in these cases, \ie the solution set is a star domain modulo permutations. Our empirical findings support this hypothesis. We propose the \ourmethod algorithm to find candidate \quotes{star models} and verify that they are indeed linearly connected to other solutions. In addition to the empirical evidence for the star domain conjecture, we present potential use cases for star models in practice, including uncertainty estimation through Bayesian model averaging, and model fusion.

\ack{
This work was supported by the German Federal Ministry of Education and Research (BMBF): T{\"u}bingen AI Center, FKZ: 01IS18039A. The authors would also like to thank Arnas Uselis and B{\'a}lint Mucs{\'a}nyi for helpful insights.
}

\bibliographystyle{abbrvnat}
\bibliography{references}

\clearpage
\appendix

\section{Implementation details}
\label{appendix:implementation_details}

In this section, we describe the setup for replicating our experimental results.

\subsection{Model training}

Our model training hyperparameters largely reflect standard practices, but we describe them here for completeness. We used NVIDIA A100 GPUs for most of our experiments. All experiments were performed on single GPUs.

\paragraph{ResNet18 on CIFAR.}
For ResNet18 models trained on CIFAR-10 and CIFAR-100, we use a batch size of 128. We normalize the data using ImageNet statistics. For data augmentation, we apply padding to the image or its horizontal mirror, and then randomly crop out a $32\times32$ region. We train for $200$ epochs using SGD with momentum $0.9$ and a weight decay of $5e-4$. The initial learning rate is $0.1$ and follows a cosine decay schedule to reach $0$ by the end of training. Star models and regular models are trained using otherwise identical hyperparameters, except that the star models use the training objective described in \cref{alg:star_training}. The differences between different models in $\sourcemodels$ and $\heldout$ come from the random seed set at the beginning of the training process. We use the following implementation for the ResNet: \href{https://github.com/kuangliu/pytorch-cifar/blob/master/models/resnet.py}{https://github.com/kuangliu/pytorch-cifar/blob/master/models/resnet.py}. Each regular model took roughly $30$ minutes to train, while the star model ($|\sourcemodels| = 50$) took roughly $6$ hours to train.

\paragraph{DenseNet-40-12 on CIFAR.}

Our DenseNet models use largely the same training settings as ResNet18. We highlight the differences here. DenseNet uses a batch size of $64$. The weight decay factor is $1e-4$, and the models are trained for $300$ epochs. The learning rate, initially $0.1$, is multiplied with $0.1$ at epochs $150$ and $225$. We use the following implementation: \href{https://github.com/andreasveit/densenet-pytorch/blob/master/densenet.py}{https://github.com/andreasveit/densenet-pytorch/blob/master/densenet.py}. Star models follow the same training recipe. Each regular model was trained for roughly $3.5$ hours. Training the star model took approximately $7$ hours.

\paragraph{VGGs on CIFAR.}
The initial learning rate is set to $0.05$ and is multiplied by $0.1$ at epochs $100$ and $150$. Other settings are identical to those used for ResNet18. We use the following implementation: \href{https://github.com/fagp/sinkhorn-rebasin/blob/main/examples/models/vgg.py}{https://github.com/fagp/sinkhorn-rebasin/blob/main/examples/models/vgg.py}. Star models follow the same recipe as regular models. It took roughly $15$ minutes to train each source model, and $35$ minutes to train a star model.

\paragraph{ResNets on ImageNet.}
For ImageNet, we use a batch size of $256$. Models are trained for $100$ epochs, using SGD with a learning rate of $0.1$ which is multiplied by $0.1$ at epochs $30, 60, 90$. The weight decay factor is $1e-4$. We use the ResNet18 implementation included in PyTorch \citep{paszke2019pytorch}. We leverage the open-source library FFCV \citep{leclerc2023ffcv} to speed up our experiments. For data augmentation, we resize the image or its horizontal mirror to $256\times256$ and randomly crop out a $224\times224$ region. Each source model took roughly $13$ hours to train, while a star model required about $2$ days.

\paragraph{Weight matching.} We use weight matching (WM) \citep{ainsworth_git_re_basin_2022}. Our implementation leverages an open-source Python package called \quotes{rebasin}: \href{https://pypi.org/project/rebasin/}{https://pypi.org/project/rebasin/}.

\paragraph{Total compute.} We estimate to have spent approximately $50-70$ days of NVIDIA A100 compute hours for the experiments in this paper (not including experiments that did not make it into the paper). 

\subsection{Calculation of loss barriers}

We use the definition in \cite{entezari_permutation_invariances_2022} to calculate loss barriers between any given pair of models (\cref{eq:loss_barrier_definition}). Since the definition in \cref{eq:loss_barrier_definition} involves an infinite search space for the maximum, we sample a finite set $T = \{t_1, t_2, \cdots, t_K\}$ of equi-spaced points and compute the maximum as

\begin{align}
    \max_{t \in T}{\genericloss{(1-t)\cdot\theta_A + t\cdot\theta_B} - ((1-t)\cdot\genericloss{\theta_A} + t\cdot\genericloss{\theta_B}}
    \label{eq:loss_barrier_empirical}
\end{align}

Our sampling of equidistant points is consistent with prior work \cite{ainsworth_git_re_basin_2022, sinkhorn_re_basin_2023}. The size of $T$ itself varies in prior work. Because of the scale of our experiments, we use $|T| = 11$, including the end-points. In \cref{appendix:interpolation_sample_sizes}, we show that this size is sufficient for obtaining statistically significant results.

\subsection{Handling batch normalization}

Batch normalization \citep{batchnorm} is integral to efficient DNN training. \cite{jordan2023repair} describe the so-called \quotes{variance collapse} problem that leads to degradation of interpolated models. As a solution, we follow \cite{ainsworth_git_re_basin_2022} and recalculate the batch statistics for each interpolated model, by performing one forward pass through the entire training set before performing evaluation.

\section{Statistical significance of our results}
\label{appendix:statistical_significance}

In this section, we validate our choices concerning the reporting of our results and demonstrate that our findings are statistically significant. 

\subsection{Sample sizes for interpolation}
\label{appendix:interpolation_sample_sizes}

An essential part of our experimental setup involves computing the loss barrier between two given networks. To achieve this, we selected a set of equally spaced points between $t=0$ and $t=1$ and evaluated the interpolated models at these points. This process is computationally intensive and becomes slower as the number of interpolation points increases. Throughout this study, we used the points $t = 0.0, 0.1, \dots, 1.0$. In this section, we demonstrate that the number of interpolation points we used does not negatively impact the significance of our results. To this end, we conducted an ablation study on CIFAR10-ResNet18 models, varying the number of interpolation points for computing the loss barrier. We present the comparison in \cref{table:interpolation_different_sample_sizes}. As we increased the number of interpolation points from $11$ to $51$, we observed a decrease of $0.007$ in the \quotes{regular-regular} barrier and an increase of $0.004$ in the \quotes{star-regular} barrier. For reference, these differences are less than the standard deviations in the corresponding observations, and are thus statistically insignificant for our final conclusions. 

\subsection{Size of held-out set}

To reduce noise in our results, we compare barriers after computing them for several model pairs. The size of the held-out set $|H|$ is usually $5$, and sometimes even $3$. Here, we confirm that this is a large enough sample size and that considering a larger set of held-out models does not change our results fundamentally. In particular, we vary the number of held-out models $|\heldout|$ and source models $|\sourcemodels|$ and obtain the corresponding mean barrier values as well as standard deviations. First, we set both $|\heldout|$ and $|\sourcemodels|$ to $3$. Then, we set $|\heldout| = |\sourcemodels| = 5$ and finally, $|\heldout| = |\sourcemodels| = 15$. In each case, we interpolate all held-out models with all source models. Hence, in the last case, we perform $225$ \quotes{regular-regular} barrier computations and $15$ \quotes{star-regualar} barrier computations. We present the results in \cref{table:interpolation_many_heldouts}. We observe that the average \quotes{regular-regular} loss barrier between two arbitrary models remains larger than $0.37$ throughout, with a standard deviation close to $0.05$. In contrast, the average \quotes{star-regular} barrier remains lower than $0.1$, with a standard deviation lower than $0.02$. None of the observed metrics or our conclusions change significantly when increasing the number of samples. This observation provides confidence that our practice of setting the number of held-out models to $3$ or $5$ provides reliable estimates while also being computationally cheaper.

\subsection{Maximum and minimum barriers}

Throughout the study, we consider mean values of \quotes{regular-regular} and \quotes{star-regular} loss barriers for comparison. Here, we additionally compare maximum and minimum barrier values for each model pair and confirm that the same trend holds, \ie \quotes{star-regular} barriers are lower than \quotes{regular-regular} barriers. We present the results in \cref{table:interpolation_many_heldouts}. We observe that as the sample size increases, the minimum barrier values go down, while the maximum barrier values go up for both \quotes{regular-regular} and \quotes{star-regular} pairs. But the minimum barrier obtained by \quotes{regular-regular} pairs is still $0.25$, which is significantly higher than even the maximum \quotes{star-regular} barrier, \ie $0.117$. This observation confirms that the star model, on average, exhibits better linear connectivity with other arbitrary models, than even the most \quotes{connected} arbitrary models exhibit between each other.

\section{Ablations}

\subsection{Test metrics}
\label{appendix:test_metrics}
\begin{table}

\renewcommand{\arraystretch}{1.2}
\setlength{\tabcolsep}{.4em}
\small
\caption{
\textbf{Empirically verifying the star domain conjecture.} 
\quotes{Regular loss} and \quotes{Star loss} indicate \textit{test} losses for regular models in $\sourcemodels{}$ and star models $\starmodel{}$, respectively. 
\quotes{Star-regular} refers to the barrier $\barrier{\starmodel{}}{\theta_h}$ between a star model and one of the heldout models in $\heldout$. 
For comparison, \quotes{Regular-regular} is the \textit{test} loss barrier $\barrier{\theta_A}{\theta_B}$ between two arbitrary models.  
We report values up to one standard deviation over several runs, except for ImageNet. 
In each case, star models exhibit significantly lower loss barriers with other models, than the corresponding average loss barrier between two regular models. This trend is consistent with our observation for training losses in \cref{table:star_model_empirical_verification}.
}
\label{table:star_model_empirical_verification_test_losses}
\vspace{.5em}
\centering
\begin{tabular}{lccccccr}
\toprule
\normalfont{Dataset} &
\normalfont{Architecture} &
\normalfont{Regular loss} &
\normalfont{Star loss} &
\normalfont{Regular-regular} &
\normalfont{Star-regular} &
\\
\midrule
CIFAR-10 &
ResNet-18 &
$0.181 \pm 0.005$ &
$0.222$ &
$0.336 \pm 0.057$ &
$0.035 \pm 0.007$ &
\\
CIFAR-10 &
ResNet-18 (ADAM) &
$0.334 \pm 0.010$ &
$0.299$ &
$1.099 \pm 0.516$ &
$0.168 \pm 0.015$ &
\\
CIFAR-10 &
VGG11 &
$0.421 \pm 0.011$ &
$0.456$ &
$0.242 \pm 0.036$ &
$0.000 \pm 0.000$ &
\\
CIFAR-10 &
VGG19 &
$0.444 \pm 0.019$ &
$0.395$ &
$0.903 \pm 0.150$ &
$0.117 \pm 0.067$ &
\\
CIFAR-10 &
DenseNet &
$0.269 \pm 0.012$ &
$0.290$ &
$4.405 \pm 0.730$ &
$1.612 \pm 0.408$ &
\\
\midrule
CIFAR-100 &
ResNet-18 &
$0.925 \pm 0.007$ &
$1.216$ &
 $2.306 \pm 0.039$ &
$0.447 \pm 0.051$ &
\\
CIFAR-100 &
DenseNet &
$1.312 \pm 0.019$ &
$1.115$ &
$5.613 \pm 0.209$ &
$3.005 \pm 0.176$ &
\\
\midrule
ImageNet-1k &
ResNet-18 &
$1.203$ &
$1.634$ &
$5.477$ &
$2.548$ &
\\
\bottomrule
\end{tabular}
\end{table}

The main discourse around mode connectivity \citep{garipov_mode_connectivity_2018} as well as convexity \citep{entezari_permutation_invariances_2022} is built around training loss and accuracy. Therefore, our empirical observations in support of the star domain conjecture also primarily use the training set. However, applications like Bayesian model averaging (\cref{subsection:bayesian_model_averaging}) require our conclusions to hold for the test set. Therefore, we also examine the veracity of our claims with respect to test loss and accuracy. We present our representative findings in \cref{table:star_model_empirical_verification_test_losses}. Similarly to the training set results (\cref{table:star_model_empirical_verification}), we observe that the \quotes{star-regular} test loss barriers are consistently lower than \quotes{regular-regular} test loss barriers. Notably, for VGG11, our star model achieves a zero test loss barrier with regular models, in comparison to a barrier of $0.24$ between two regular models. Overall, our observations indicate that the star domain conjecture holds for both training and test losses.

\begin{table}
\renewcommand{\arraystretch}{1.2}
\setlength{\tabcolsep}{.4em}
\small
\caption{
\textbf{Different sample sizes for computing loss barriers.}
We report the \quotes{regular-regular} and \quotes{star-regular} loss barriers computed using \cref{eq:loss_barrier_empirical}, with different sizes of the set $T$ of interpolation points. 
Using $51$ interpolation points instead of $11$ does not lead to a significant change in the computed loss barriers.
}
\label{table:interpolation_different_sample_sizes}
\vspace{.5em}
\centering
\begin{tabular}{lcccr}
\toprule
\normalfont{Sample size} &
\normalfont{Regular-regular} &
\normalfont{Star-regular} &
\\
\midrule
$11$ &
$0.376 \pm 0.055$ &
$0.090 \pm 0.007$ &
\\
\midrule
$21$ &
$0.369 \pm 0.053$ &
$0.089 \pm 0.006$ &
\\
\midrule
$51$ &
$0.369 \pm 0.053$ &
$0.094 \pm 0.008$ &
\\
\bottomrule
\end{tabular}
\end{table}

\begin{table}
\renewcommand{\arraystretch}{1.2}
\setlength{\tabcolsep}{.4em}
\small
\caption{
\textbf{Computing average loss barriers over different sizes of the sets of source models $\sourcemodels$ and heldout models $\heldout$.}
We compute minimum, average and maximum \quotes{star-regular} barriers (\quotes{SR-min}, \quotes{SR-avg.} and \quotes{SR-max.} respectively) over varying numbers of heldout models $\heldout$. For comparison, we compute minimum, average and maximum \quotes{regular-regular} barriers (\quotes{RR-min}, \quotes{RR-avg.} and \quotes{RR-max.} respectively). Increasing the number of held-out models from $3$ to $15$ does not significantly change the observed trend. 
}
\label{table:interpolation_many_heldouts}
\vspace{.5em}
\centering
\begin{tabular}{lcccccccccr}
\toprule
\normalfont{Sample size} &
\normalfont{RR-min.} &
\normalfont{SR-min.} &
\normalfont{RR-avg.} &
\normalfont{SR-avg.} &
\normalfont{RR-max.} &
\normalfont{SR-max.} &
\\
\midrule
3 &
$0.300$ &
$0.082$ &
$0.376 \pm 0.055$ &
$0.090 \pm 0.007$ &
$0.445$ &
$0.095$ &
\\
5 &
$0.292$ &
$0.069$ &
$0.381 \pm 0.059$ &
$0.077 \pm 0.007$ &
$0.530$ &
$0.084$ &
\\
15 &
$0.255$ &
$0.071$ &
$0.382 \pm 0.046$ &
$0.093 \pm 0.014$ &
$0.540$ &
$0.117$ &
\\
\bottomrule
\end{tabular}
\end{table}

\begin{figure*}
    \centering
    \includegraphics[width=\linewidth]{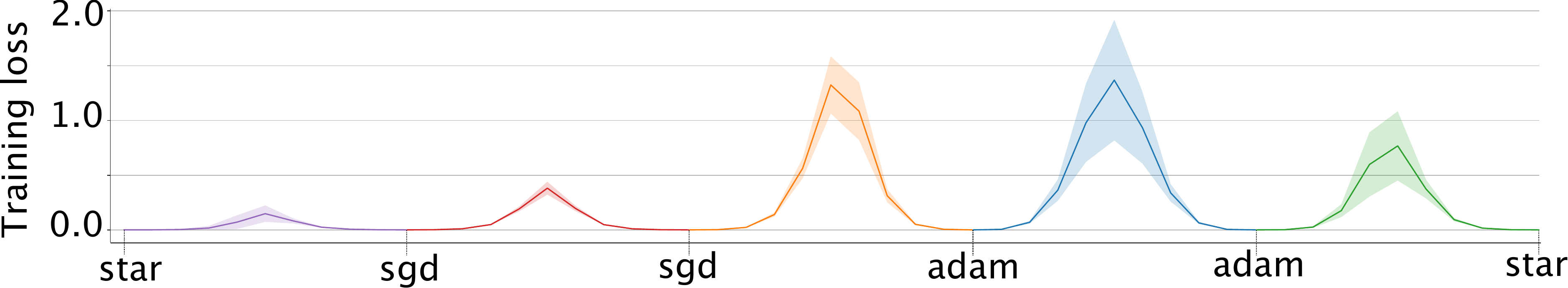}
    \caption{\textbf{Training loss landscape across SGD models, Adam models, and SGD-induced star models.} We plot test loss across different types of solutions in $\solutionset$. Our star model $\theta^\star$ (``star'' in the plot) is constructed from a set of SGD-trained models $Z$. 
    We note that the star model is well-connected with SGD solutions. There remains a loss barrier between the star model and Adam solutions, but it is significantly lower than the barrier among Adam solutions. 
    }
    \label{figure:barriers_diverse_models}
\end{figure*}

\subsection{Connectivity to Adam Solutions}
Throughout the study, we have largely focused on the SGD-trained solutions and the star model built from SGD-trained source models. 
Here, we examine the connectivity between SGD solutions, SGD-induced star models and Adam-trained solutions.
\cref{figure:barriers_diverse_models} shows the loss landscape across different types of solutions.
We observe, as before, the loss barrier between SGD solutions and our star model $\starmodel{}$ is nearly non-existent, while the SGD solutions are generally not linearly connected.
Our star model $\starmodel{}$ shows less connectivity with Adam solutions (the curve between ``adam'' and ``star-sgd'') than with SGD solutions.
However, we note that the loss barrier is significantly lower than for the linear interpolations between pairs of Adam solutions (the curve between ``adam'' and ``adam'').
Based on the observation, we conclude that while the scope of our conjecture remains within SGD-trained solutions, there are hints that our star model shows enhanced connectivity with other types of solution subsets.

\subsection{Comparison with Sinkhorn-rebasin} 
\citet{sinkhorn_re_basin_2023} introduced a novel permutation-finding algorithm \viz \textit{Sinkhorn-rebasin} aimed at reducing the loss barriers between two arbitrary models. When using the data-free setting, the method \cite{sinkhorn_re_basin_2023} can be considered a differentiable form of weight matching \cite{ainsworth_git_re_basin_2022}. The authors notably demonstrate that their method performs better than weight matching on average, albeit it only partially eliminates loss barriers between two given models. However, Sinkhorn-rebasin does not yet support networks with skip connections, making it unsuitable for our experiments involving ResNets and DenseNets. Additionally, Sinkhorn-rebasin requires hyperparameter tuning (such as optimizer and learning rate), which could introduce confounding factors into our experiments.

\begin{figure*}[t]
    \centering
    \small
    \setlength{\tabcolsep}{.1em}
    \begin{tabular}{cc}
        \includegraphics[width=.48\linewidth]{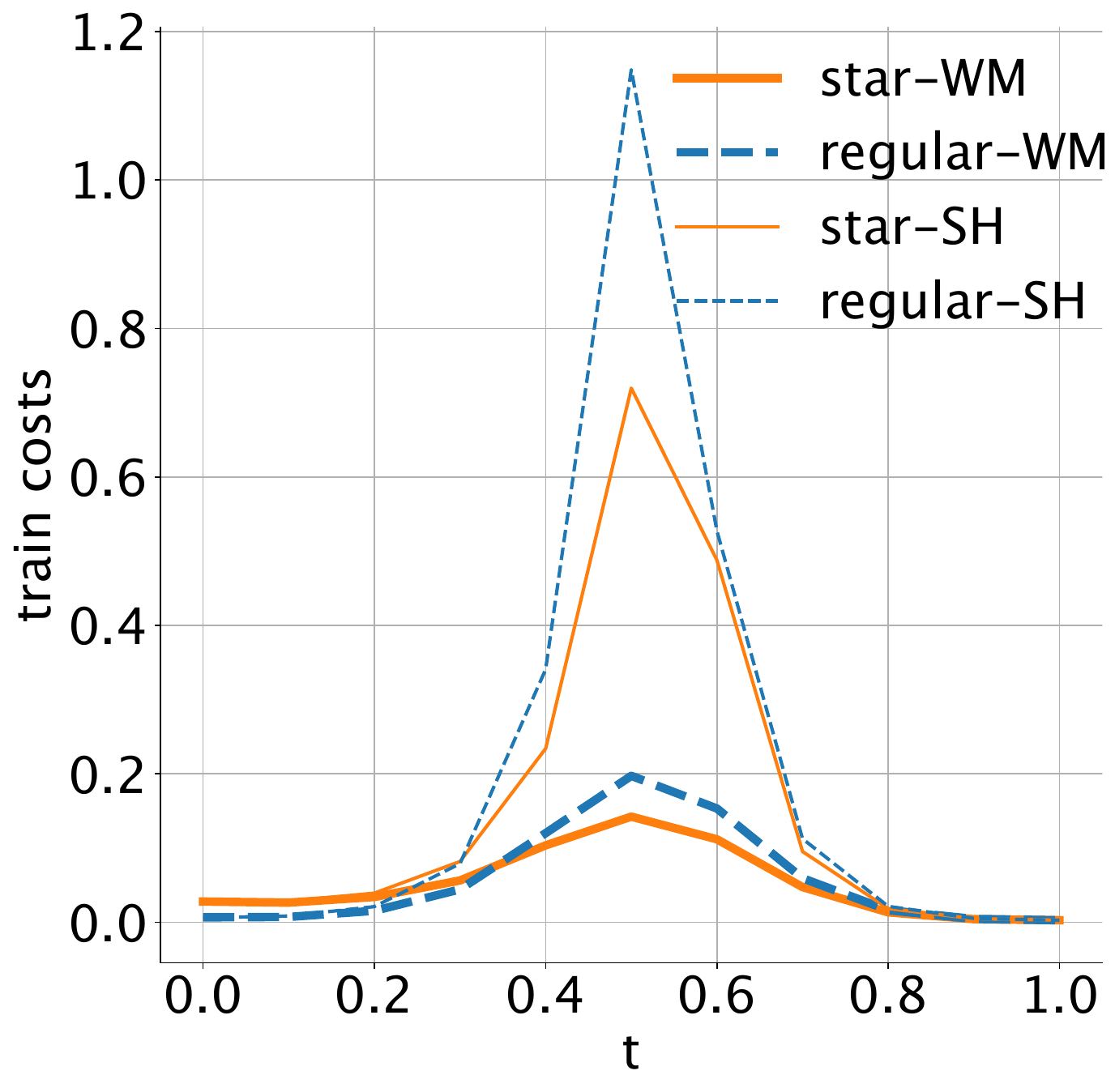} &
        \includegraphics[width=.48\linewidth]{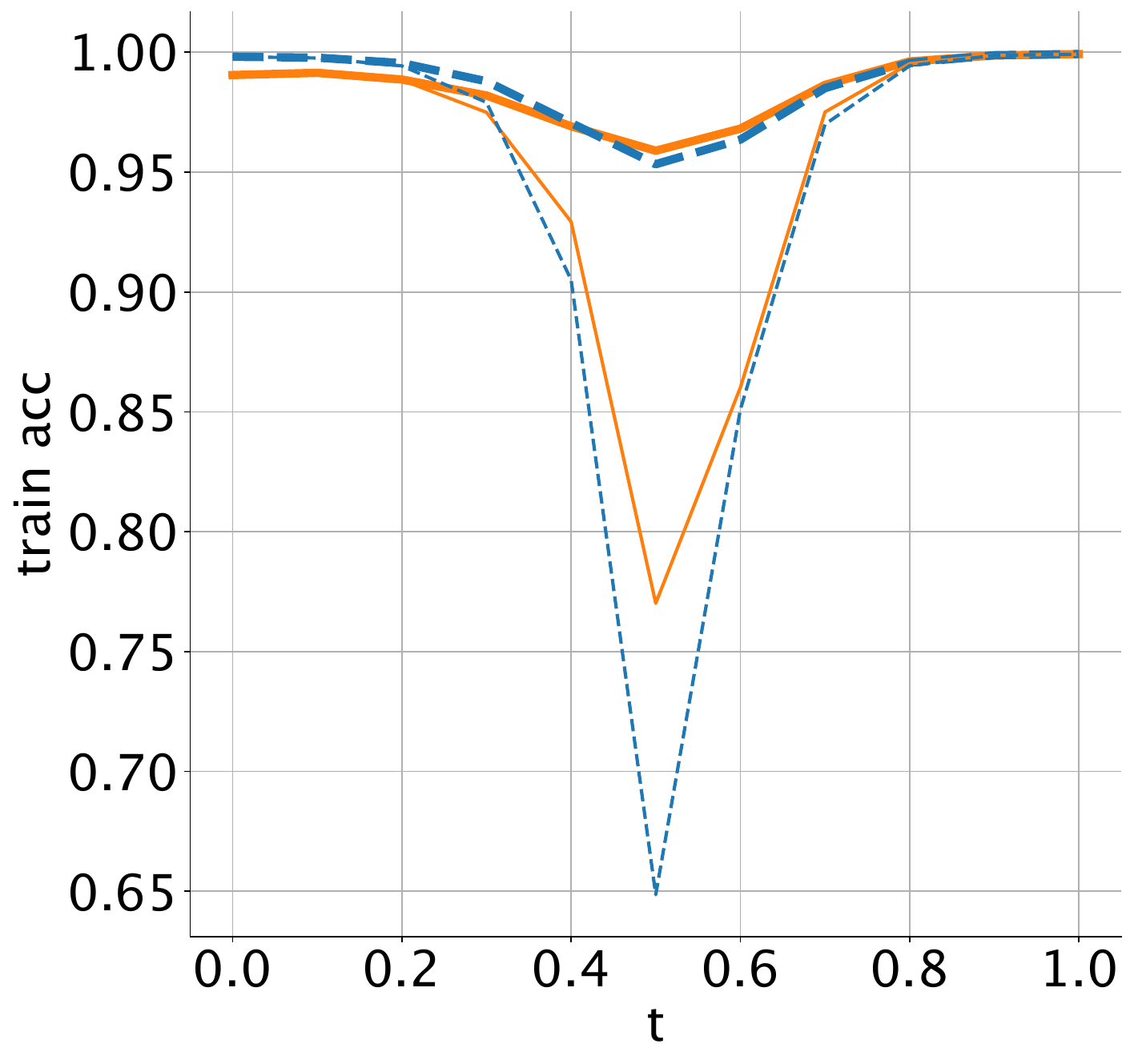} 
        \\
        \includegraphics[width=.48\linewidth]{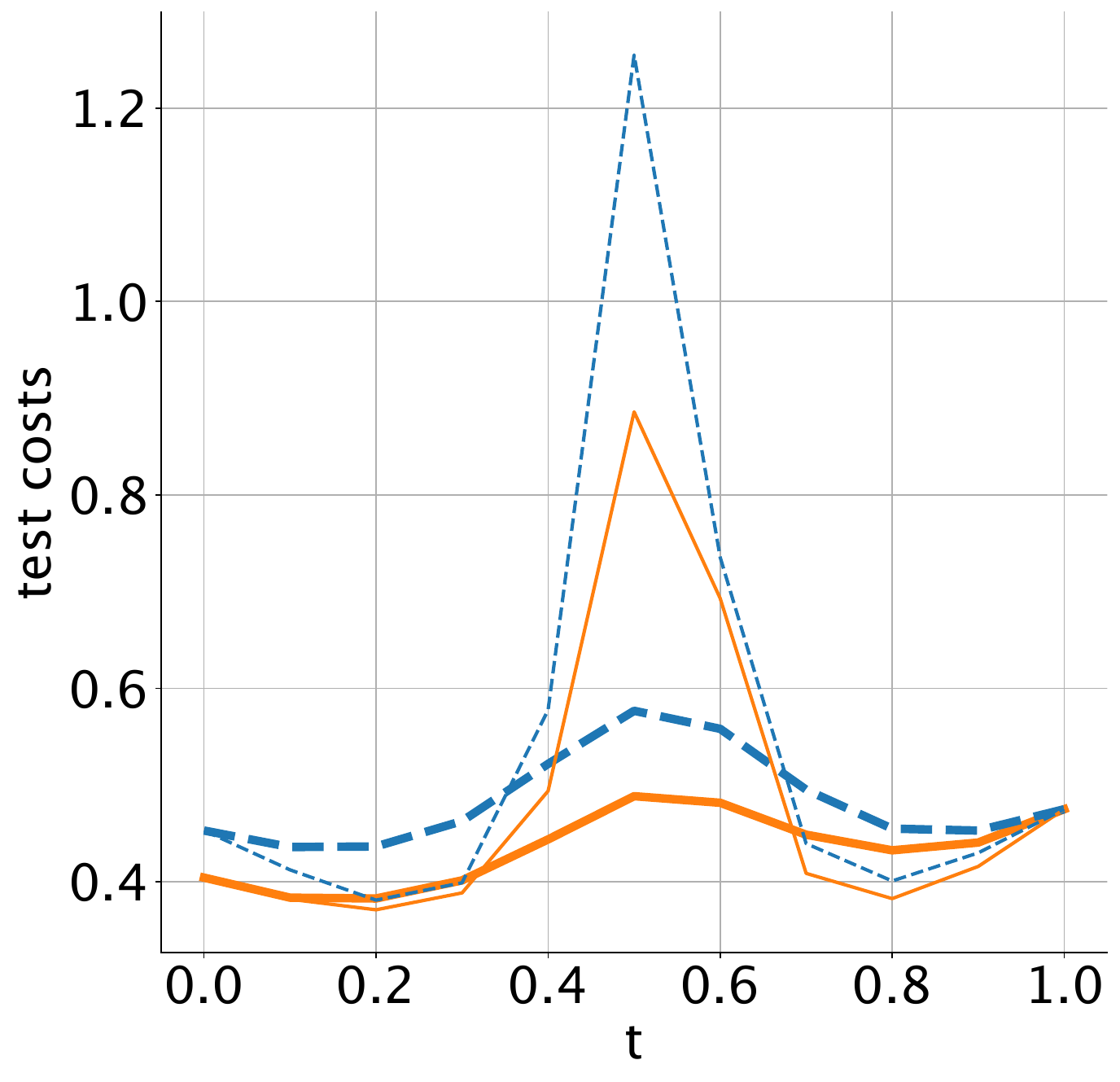} &
        \includegraphics[width=.48\linewidth]{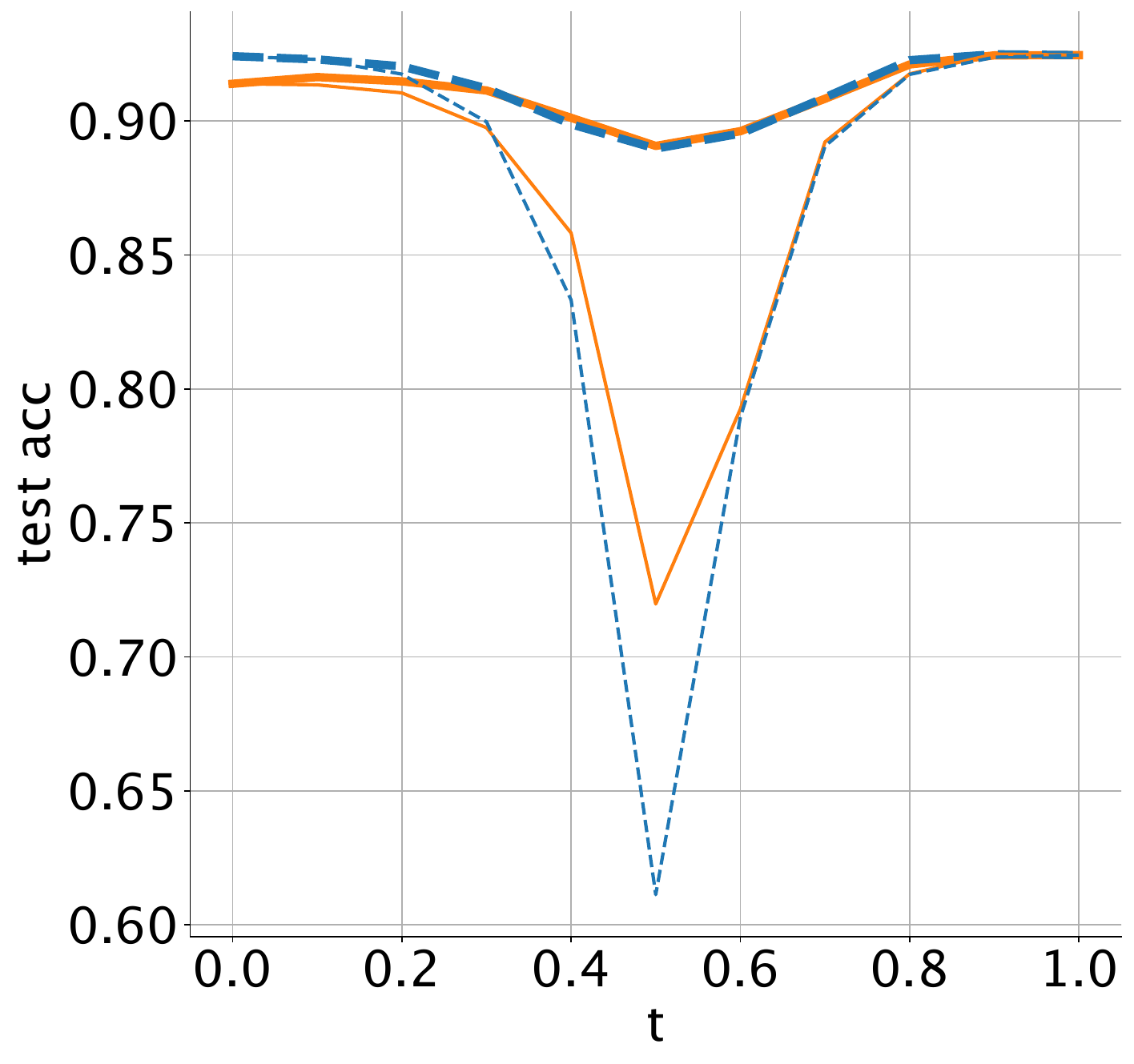} 
        \\
    \end{tabular}
    \caption{\textbf{Effect of using Sinkhorn-rebasin instead of weight matching.}
    We compare \quotes{star-regular} barriers against \quotes{regular-regular} barriers after fixing the permutation algorithm used to match the weights of the two models being interpolated. \starcolor{star-WM} and \regcolor{regular-WM} refer to the barriers after applying weight matching \cite{ainsworth_git_re_basin_2022}.
    Similarly, \starcolor{star-SH} and \regcolor{regular-SH} refer to barriers after applying Sinkhorn-rebasin \cite{sinkhorn_re_basin_2023}. 
    While vanilla WM outperforms SH in this case, a better hyperparameter choice may eventually cause SH to outperform WM. Nevertheless, the findings in this investigation further support our conclusion that star models are well-connected to other regular models, in comparison to how well the regular models are connected amongst themselves.
    }
    \label{figure:sinkhorn_comparison}
\end{figure*}

The primary objective of this paper is to investigate the empirical validity of our star domain conjecture. We find weight matching sufficient for this purpose. However, our empirical verification is based upon comparing \quotes{regular-regular} barriers, \ie loss barriers between two arbitrary solutions, and \quotes{star-regular} barriers, \ie loss barriers between the star model and other arbitrary solutions. it is important to verify how much Sinkhorn-rebasin can further improve these \quotes{regular-regular} loss barriers. To this end, we compare \quotes{star-regular} barriers \quotes{regular-regular} barriers after applying Sinkhorn-rebasin (SH) instead of Weight Matching. We use VGG19 models with batch normalization, trained on CIFAR-10, for this purpose. First, we perform a hyperparameter search on the learning rate for the permutation-finding algorithm \cite{sinkhorn_re_basin_2023}, using $\mathcal{C}_{L2}$ distance (as described in \cite{sinkhorn_re_basin_2023}) as the optimization objective. Our search space is the set ${0.01, 0.1, 1.0, 10.0, 20.0, 30.0, 40.0, 50.0, 60.0, 80.0, 100.0, 120.0, 150.0, 200.0}$. We observe the best \quotes{regular-regular} loss barrier between two pretrained models from \cite{sinkhorn_re_basin_2023}, at a learning rate of $150.0$. This loss barrier forms the reference point \quotes{regular-regular} in our comparison. Next, we train a star model using our own source models, and then compute the \quotes{star-regular} barrier with one of the pre-trained models from \cite{sinkhorn_re_basin_2023}. The results are presented in \cref{figure:sinkhorn_comparison}. 

The \quotes{regular-regular} barriers obtained in our experiment are comparable to those reported in Figure 6 of \cite{sinkhorn_re_basin_2023}. We observe that in this particular case, Sinkhorn-rebasin exhibits higher loss barriers than weight matching, although it might be possible to reduce this barrier further with a different set of hyperparameters. Nevertheless, \quotes{star-regular} barriers remain lower than \quotes{regular-regular} barriers in both cases. Future work may look more closely into the effect of using different permutation algorithms.

\subsection{Different sampling schemes for Monte-Carlo pptimization}
\label{appendix:different_sampling_schemes}

While training our star models using Starlight (\cref{alg:star_training}), we sample the interpolation factor $t$ from $\text{Unif} [0, 1]$. However, empirical results show that loss barriers are typically achieved close to the center of the interpolation line ($t = 0.5$). This raises the question: would Starlight be more efficient if the sampling scheme placed more weight toward the center of the interpolation line? To test this, we run ablations with different sampling schemes. 

\begin{itemize}
    \item \textbf{Uniform (used in main paper)}: $t \sim \text{Unif}[0, 1]$.
    \item Beta: $t \sim \text{Beta}(2, 2)$. We sample more around $t=0.5$.
    \item Constant 0.5: $t = 0.5$.
\end{itemize}

We show the results in \cref{table:starlight_sampling_schemes}. We observe that both the Uniform and Beta sampling schemes obtain star models with identical training loss ($0.001$). Beta achieves a slightly better loss barrier ($0.069$) than Uniform ($0.084$), while the difference is not particularly significant in this case. Constant sampling obtains a much worse star model in terms of training loss ($0.08$). These findings suggest that \ourmethod can potentially benefit from better sampling schemes for $t$ in future work.

\begin{table}
\renewcommand{\arraystretch}{1.2}
\setlength{\tabcolsep}{.4em}
\small
\caption{
\textbf{Different sampling schemes for \ourmethod.}
We report the \quotes{star-regular} and \quotes{regular-regular} barriers for each sampling case; averages were computed over three different runs using different random seeds. 
While sampling from a beta distribution \quotes{Beta} performs slightly better, the difference is too small to be statistically significant.
}
\label{table:starlight_sampling_schemes}
\vspace{.5em}
\centering
\begin{tabular}{lcccccr}
\toprule
\normalfont{Sampling scheme} &
\normalfont{Regular loss} &
\normalfont{Star loss} &
\normalfont{Regular-regular} &
\normalfont{Star-regular} &
\\
\midrule
$\text{Unif} [0, 1]$ &
$0.001 \pm 0.000$ &
$0.001 \pm 0.000$ &
$0.383 \pm 0.056$ &
 $0.084 \pm 0.012$ &
\\
$\text{Beta} (2, 2)$ &
-- do -- &
$0.001 \pm 0.000$ &
-- do -- &
$0.069 \pm 0.010$ &
\\
$\text{Constant} 0.5$ &
-- do -- &
$0.082 \pm 0.004$ &
-- do -- &
$0.018 \pm 0.009$ &
\\
\bottomrule
\end{tabular}
\end{table}

\section{Additional Results}
\label{appendix:additional_results}

\begin{figure*}[t]
    \centering
    \small
    \setlength{\tabcolsep}{.1em}
    \begin{tabular}{ccc|cc}
        \rotatebox[origin=c]{90}{\small CIFAR10\hspace{-5em}} & 
        \includegraphics[width=.24\linewidth]{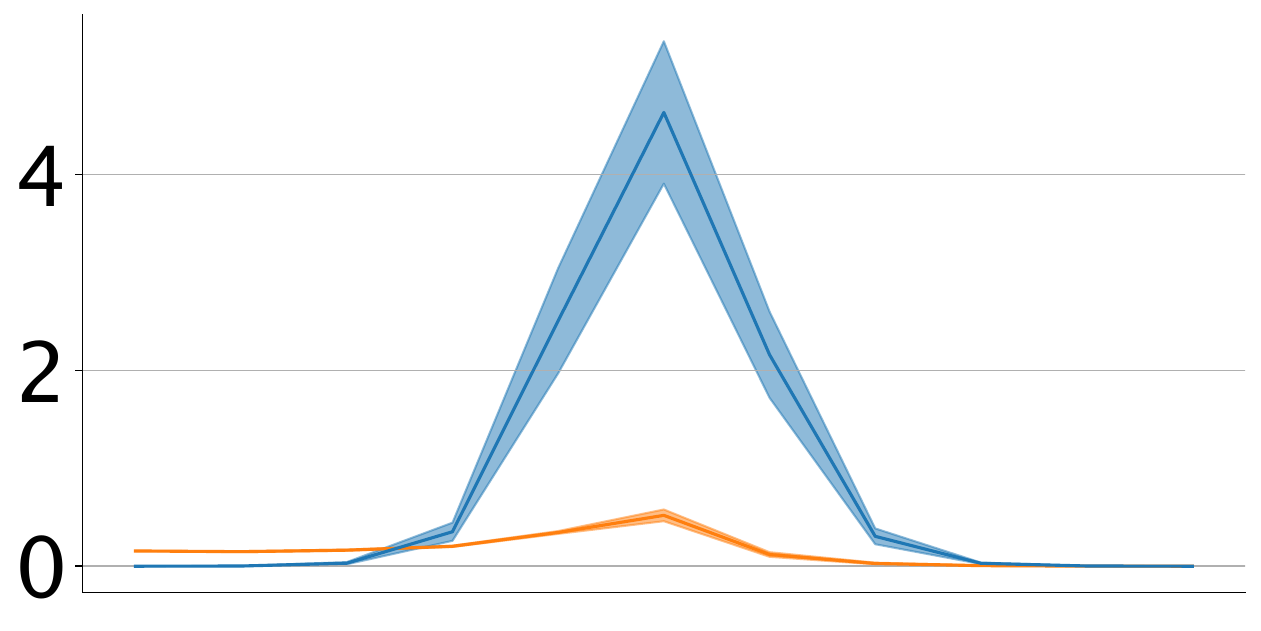} &
        \includegraphics[width=.24\linewidth]{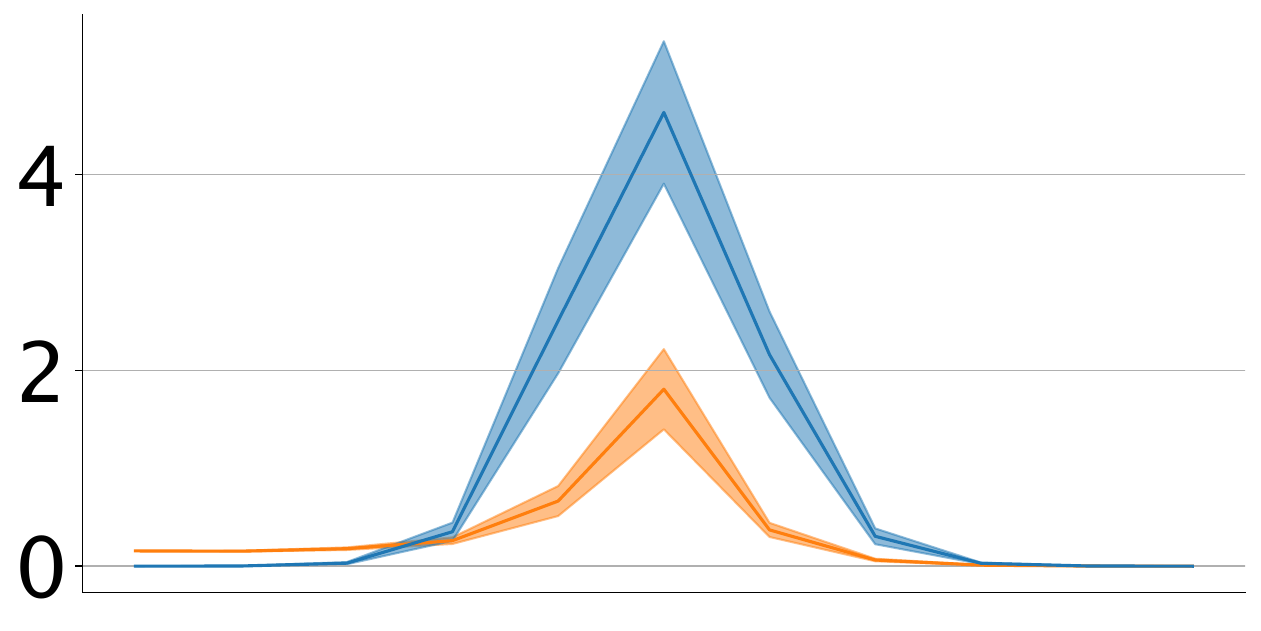} &
        \includegraphics[width=.24\linewidth]{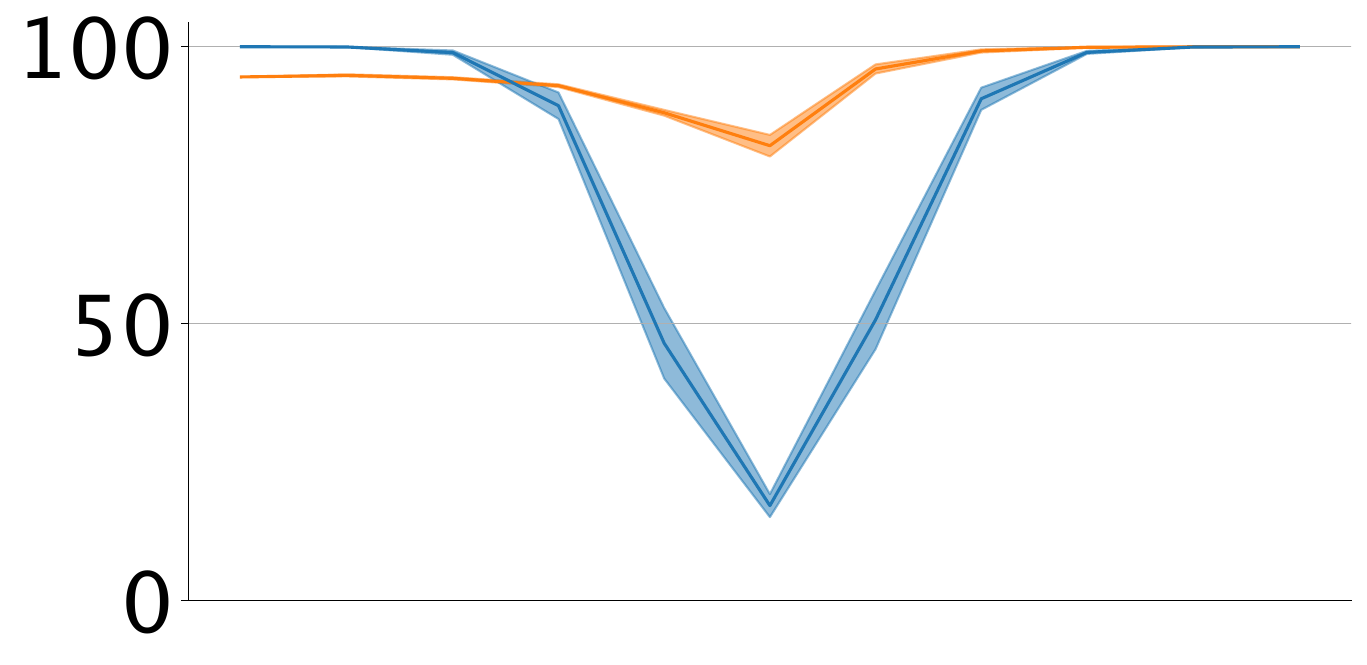} &
        \includegraphics[width=.24\linewidth]{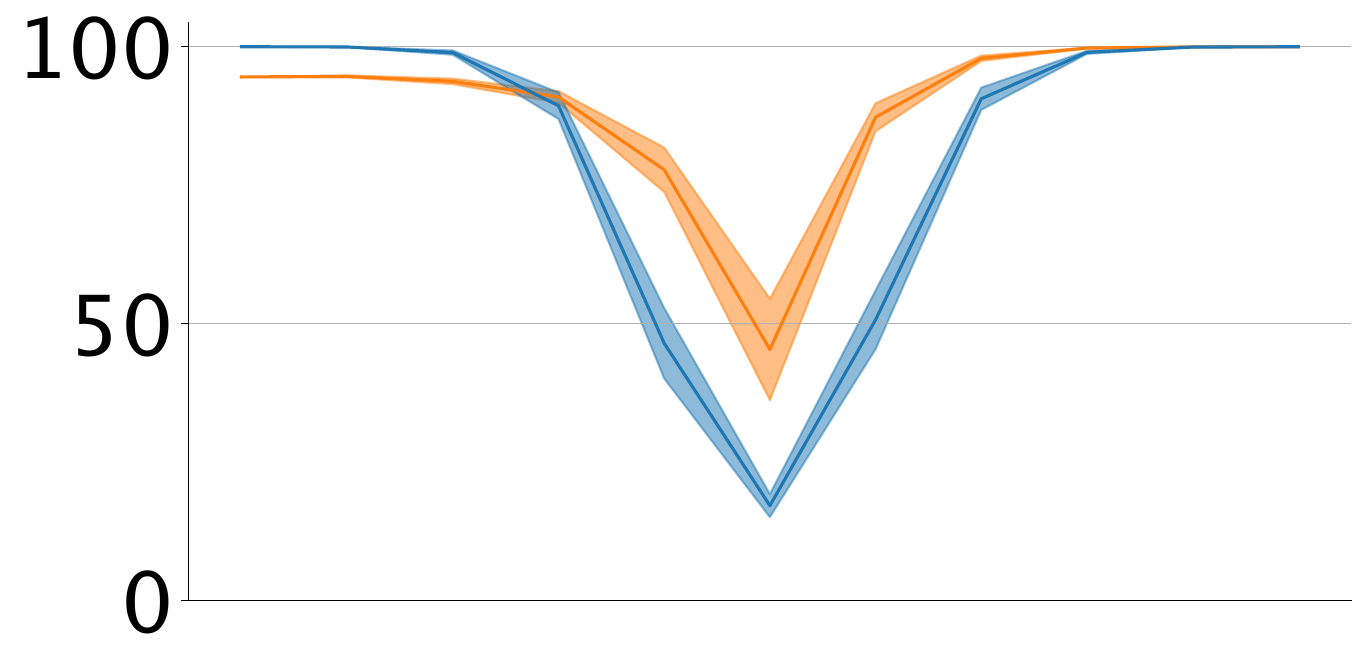}
        \\
        \rotatebox[origin=c]{90}{\small CIFAR100\hspace{-6em}} & 
        \includegraphics[width=.24\linewidth]{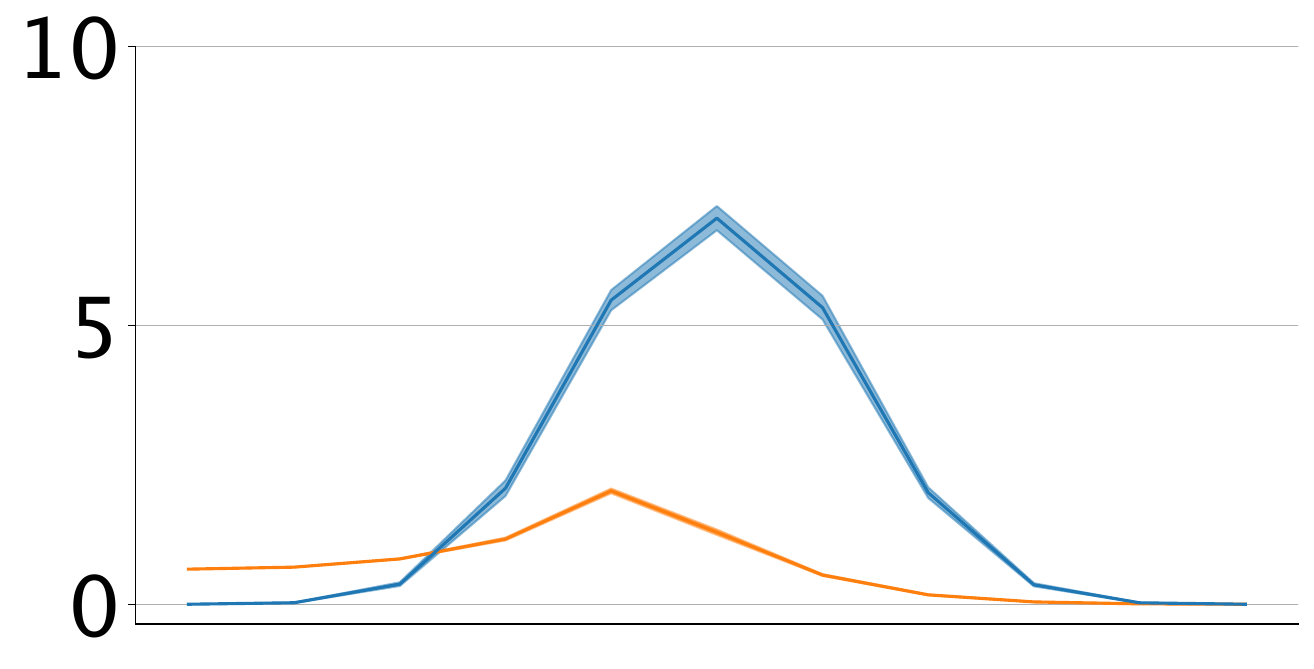} &
        \includegraphics[width=.24\linewidth]{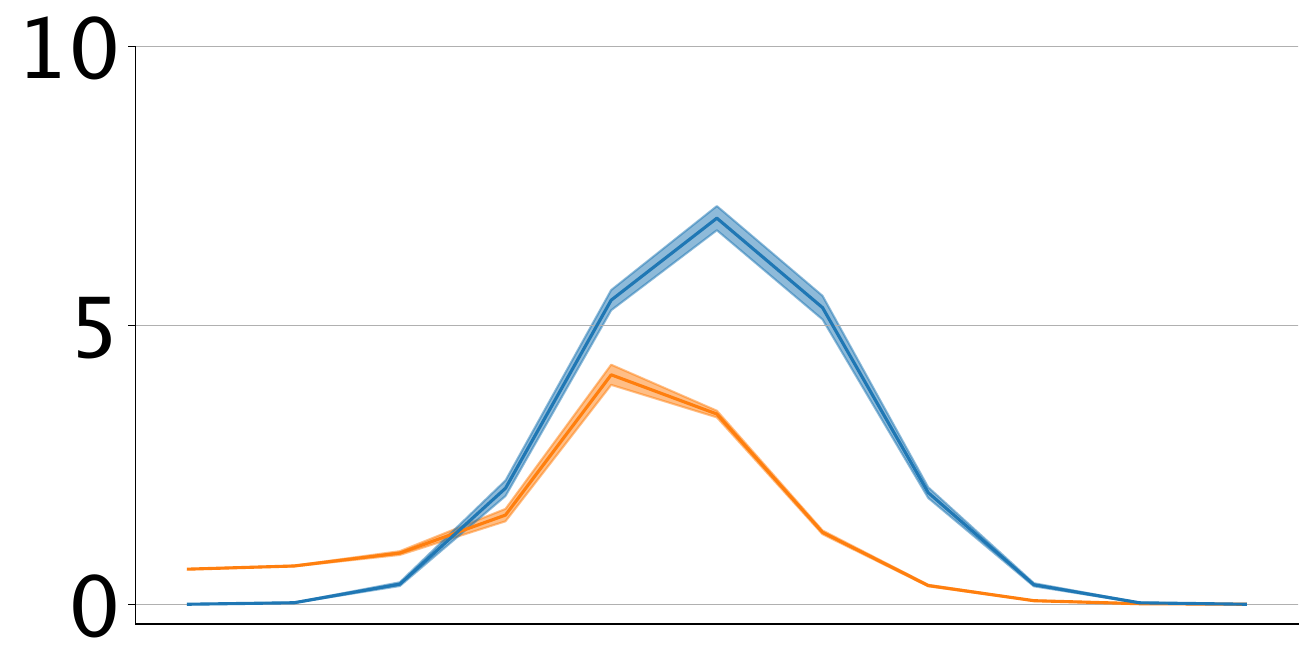} &
        \includegraphics[width=.24\linewidth]{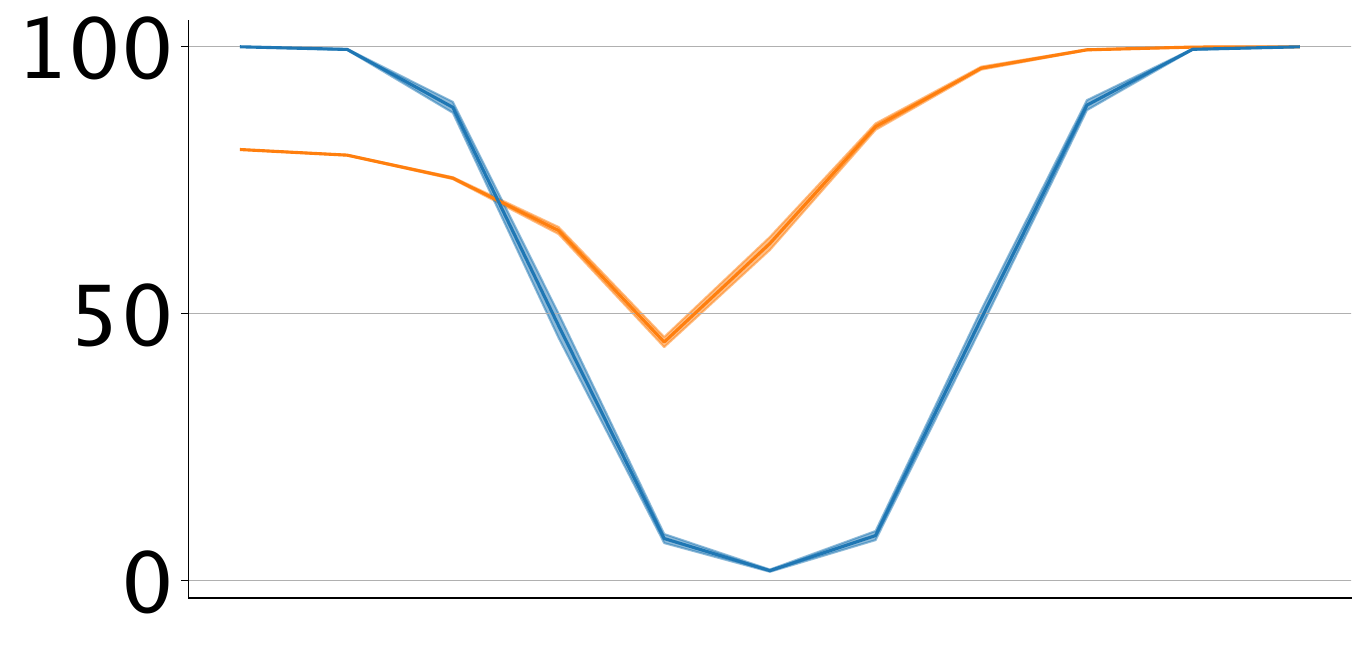} &
        \includegraphics[width=.24\linewidth]{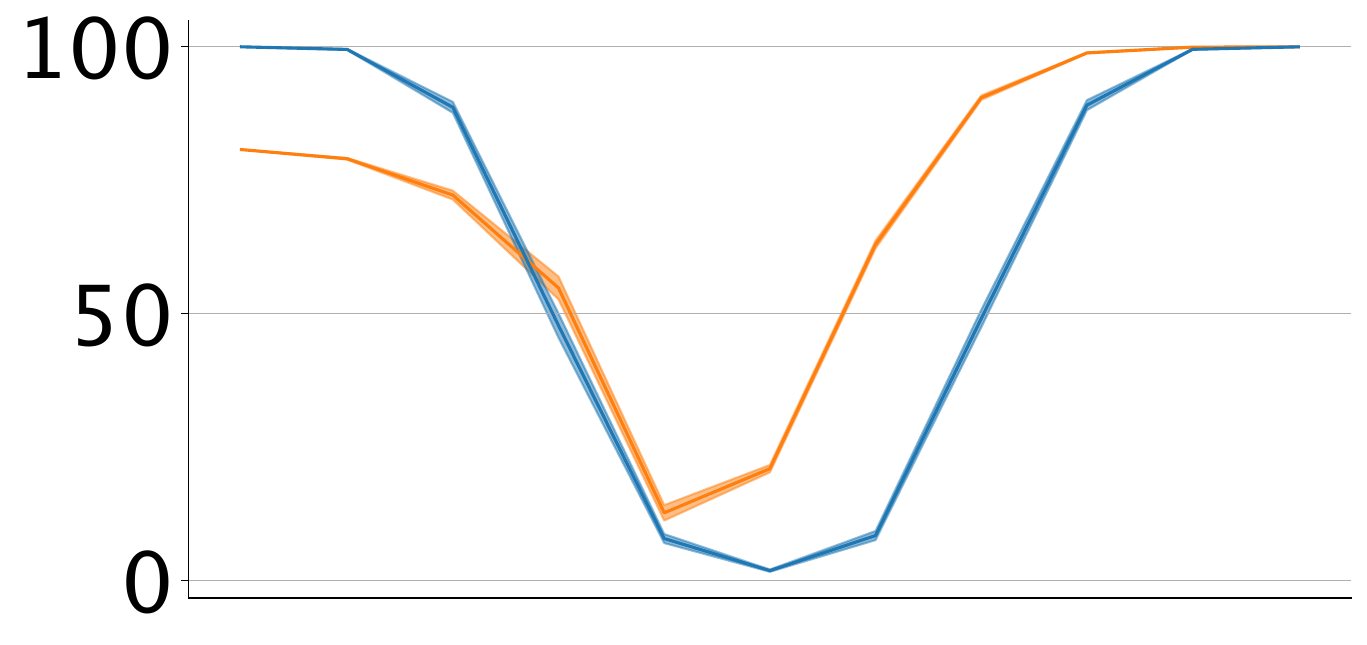}
        \\
        & star $\theta^\star$ - source $Z$ & star $\theta^\star$ - heldout $H$ & star $\theta^\star$ - source $Z$ & star $\theta^\star$ - heldout $H$ \\
        & \multicolumn{2}{c}{Train loss}
        & \multicolumn{2}{c}{Train accuracy}
    \end{tabular}
    \caption{\textbf{Loss barriers for DenseNet-40-12 star models on CIFAR}.
    We plot training loss and accuracy curves obtained upon interpolation between \starcolor{star-regular} and \regcolor{regular-regular} models pairs. 
    Star-regular loss barriers continue to be lower than regular-regular barriers, as observed in \cref{figure:cifar_resnet_interpolation_plots}.
    }
    \label{figure:densenet_cifar_interpolation_plots}
\end{figure*}

\begin{figure*}[t]
    \centering
    \small
    \setlength{\tabcolsep}{.1em}
    \begin{tabular}{ccc|cc}
        \rotatebox[origin=c]{90}{\small VGG11\hspace{-4em}} & 
        \includegraphics[width=.24\linewidth]{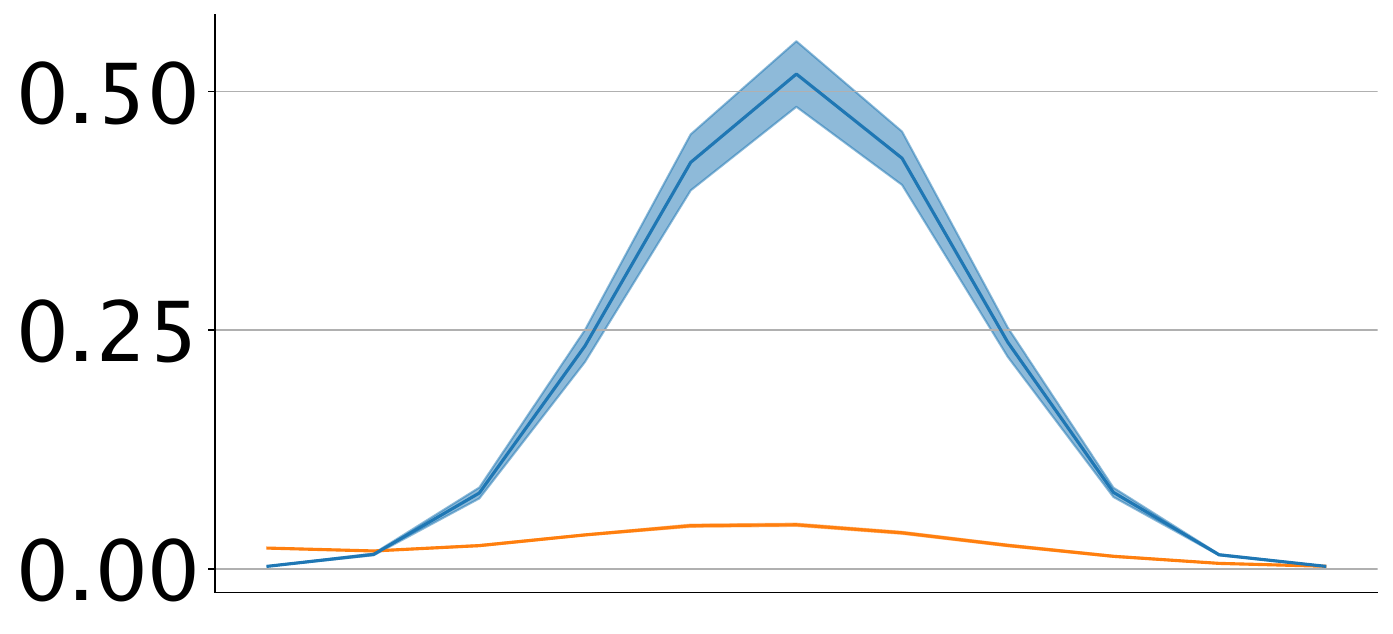} &
        \includegraphics[width=.24\linewidth]{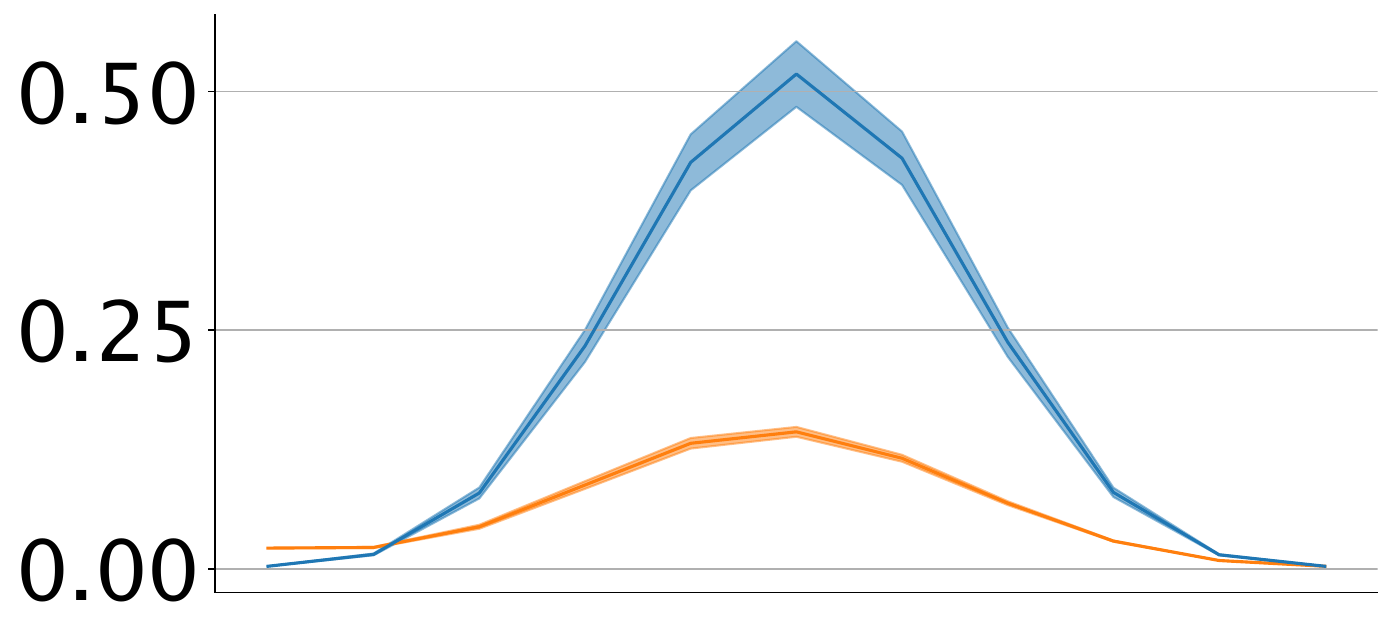} &
        \includegraphics[width=.24\linewidth]{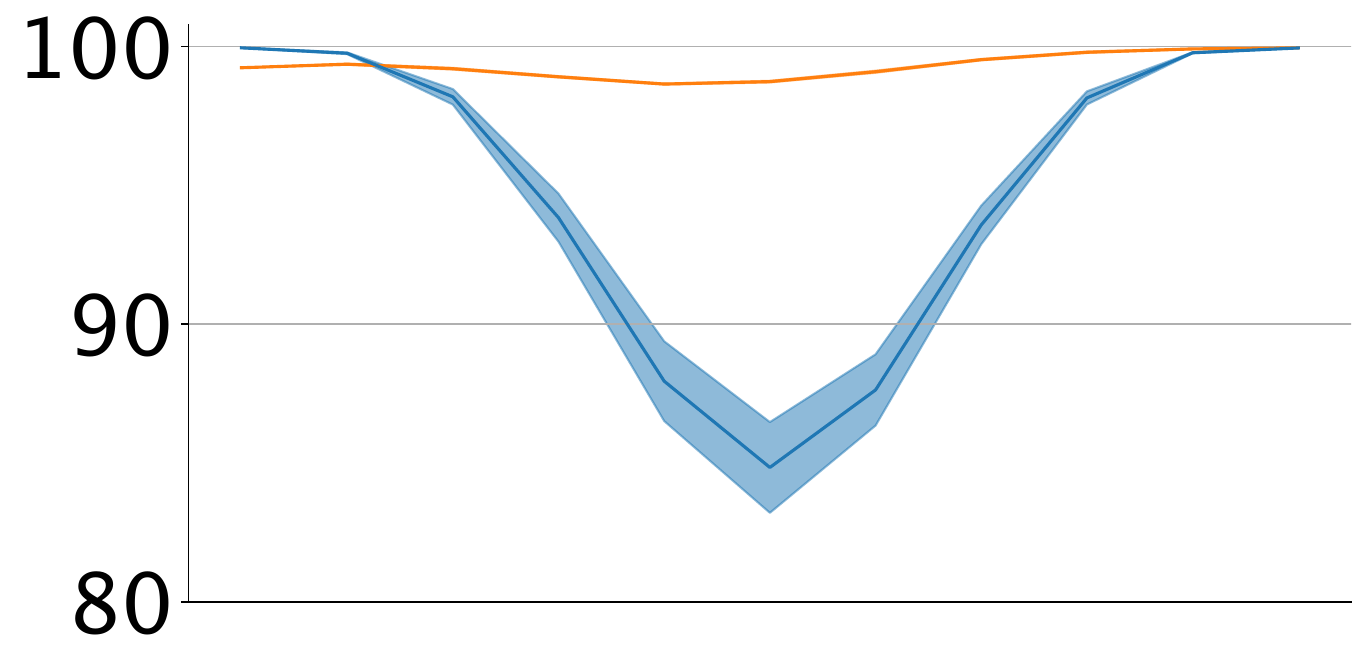} &
        \includegraphics[width=.24\linewidth]{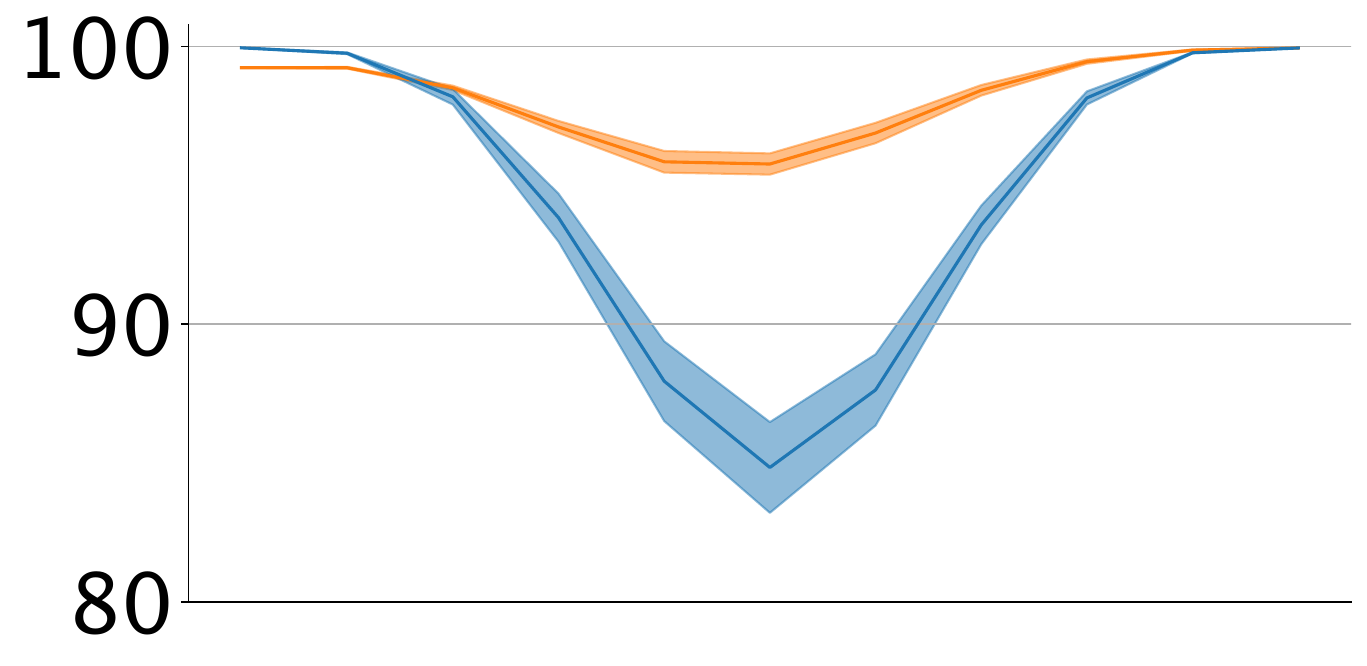}
        \\
        \rotatebox[origin=c]{90}{\small VGG19\hspace{-4em}} & 
        \includegraphics[width=.24\linewidth]{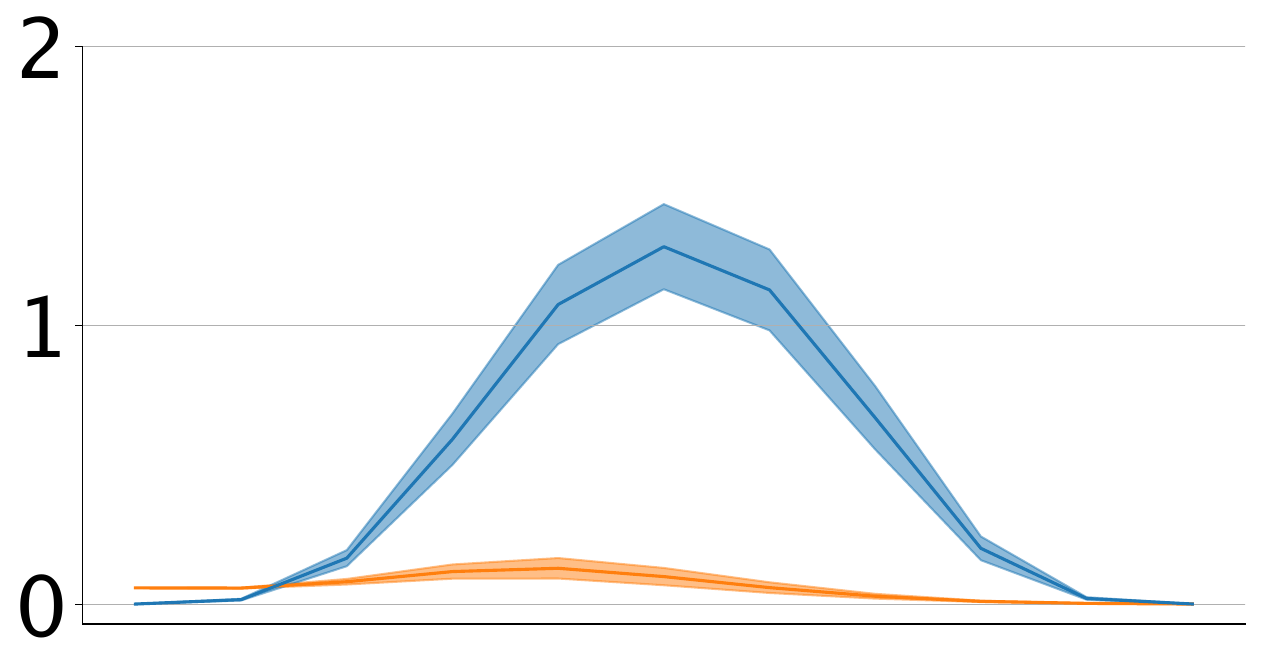} &
        \includegraphics[width=.24\linewidth]{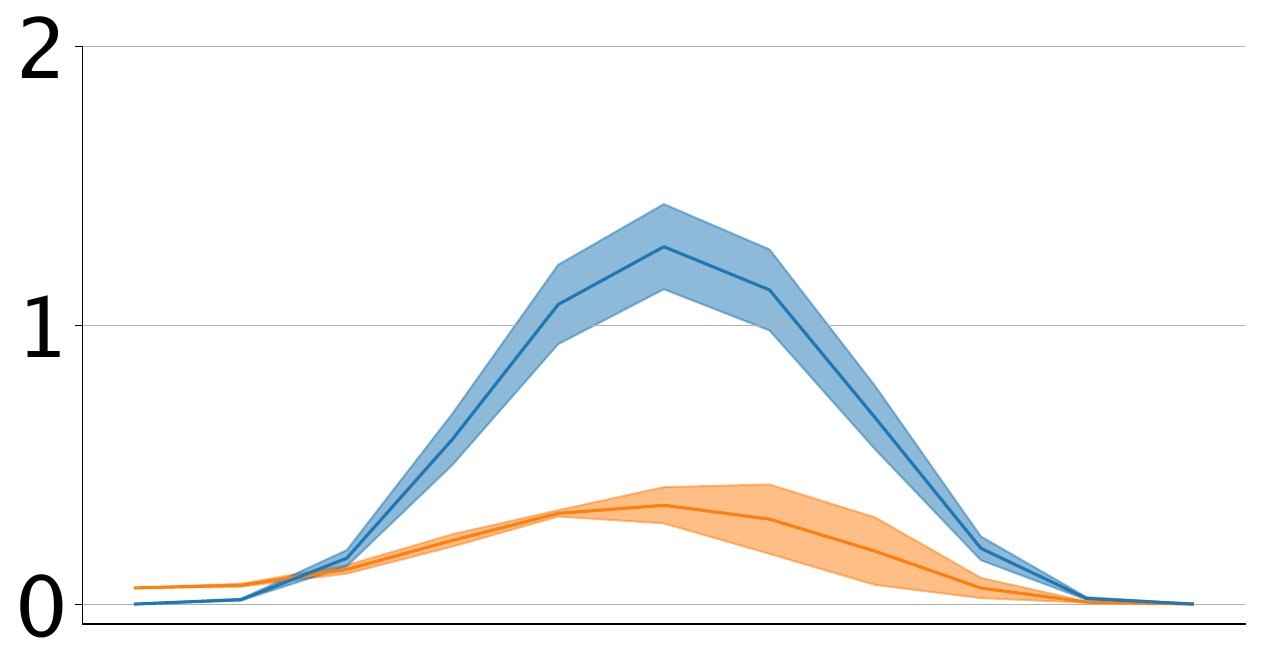} &
        \includegraphics[width=.24\linewidth]{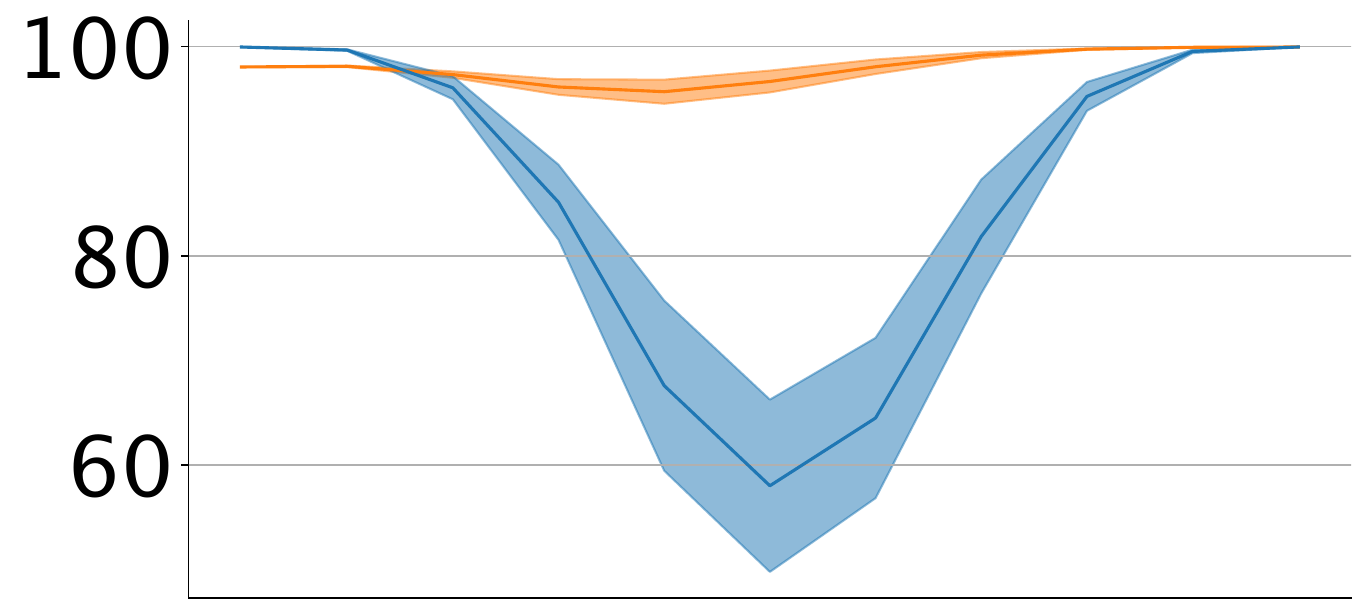} &
        \includegraphics[width=.24\linewidth]{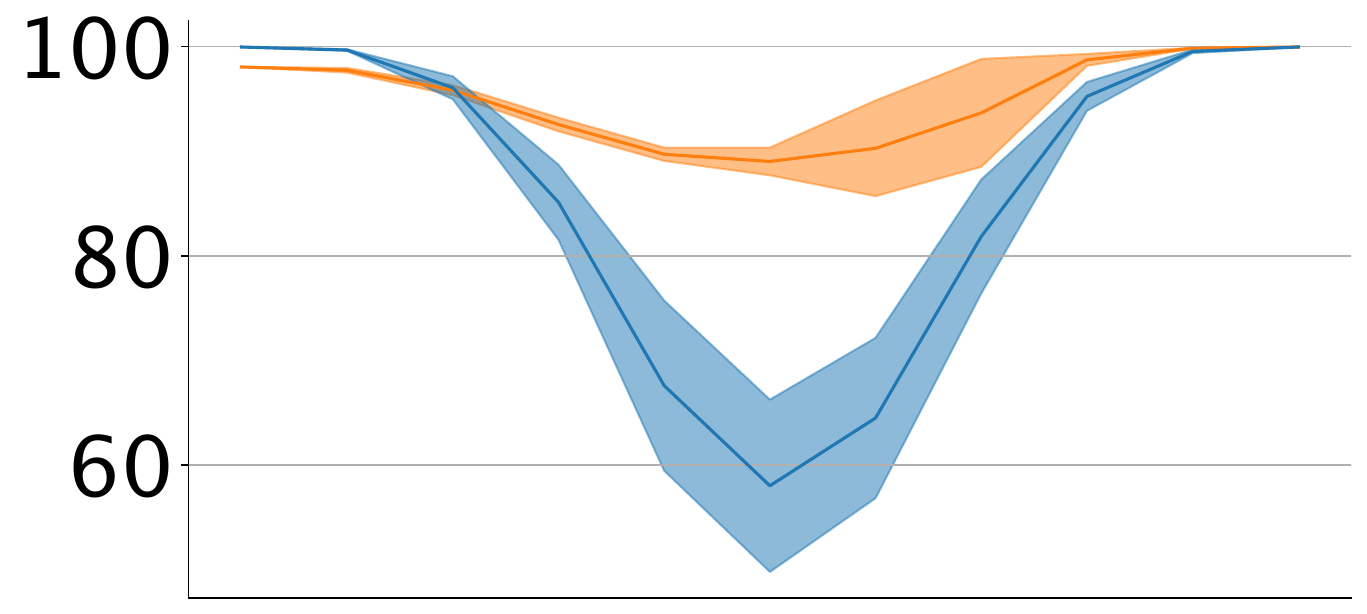}
        \\
        & star $\theta^\star$ - source $Z$ & star $\theta^\star$ - heldout $H$ & star $\theta^\star$ - source $Z$ & star $\theta^\star$ - heldout $H$ \\
        & \multicolumn{2}{c}{Train loss}
        & \multicolumn{2}{c}{Train accuracy}
    \end{tabular}
    \caption{\textbf{Loss barriers for VGG star models trained on CIFAR-10}. 
    Training loss and accuracy curves obtained upon interpolation between \starcolor{star-regular} and \regcolor{regular-regular} model pairs. We observe the same trend as in \cref{figure:cifar_resnet_interpolation_plots}.
    }
    \label{figure:vgg_cifar_interpolation_plots}
\end{figure*}

\begin{figure*}[t]
    \centering
    \small
    \setlength{\tabcolsep}{.1em}
    \begin{tabular}{cc|cc}
        \includegraphics[width=.24\linewidth]{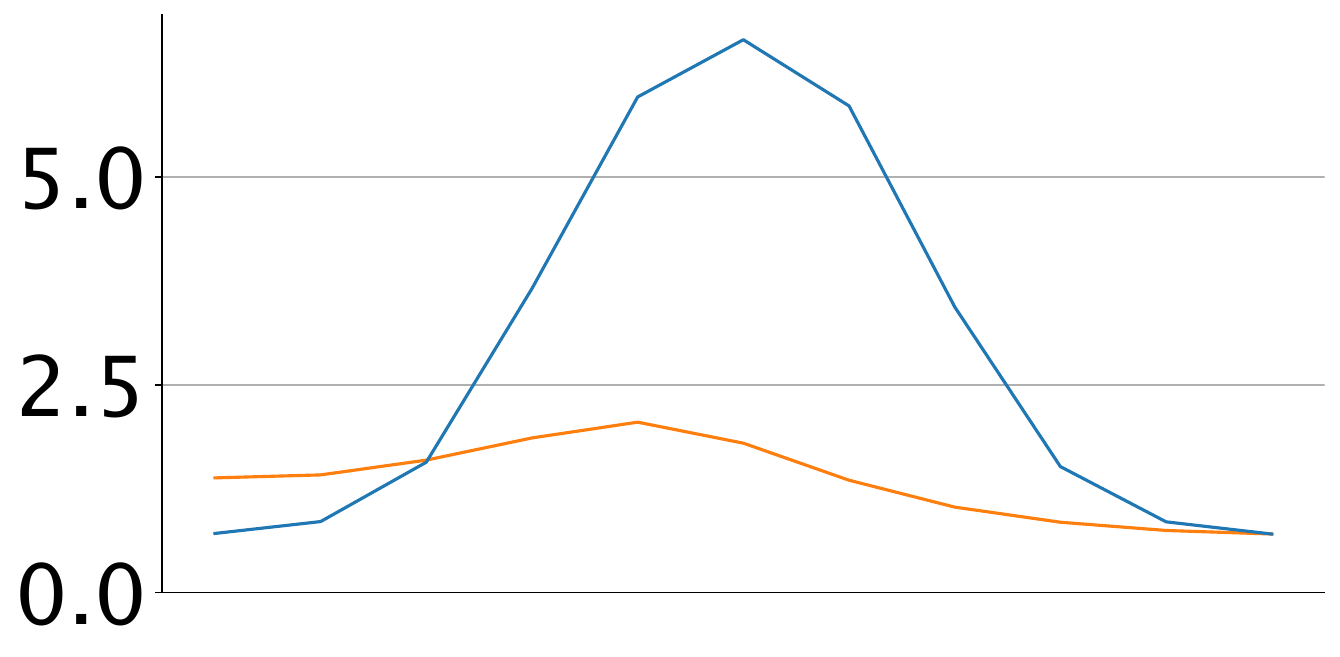} &
        \includegraphics[width=.24\linewidth]{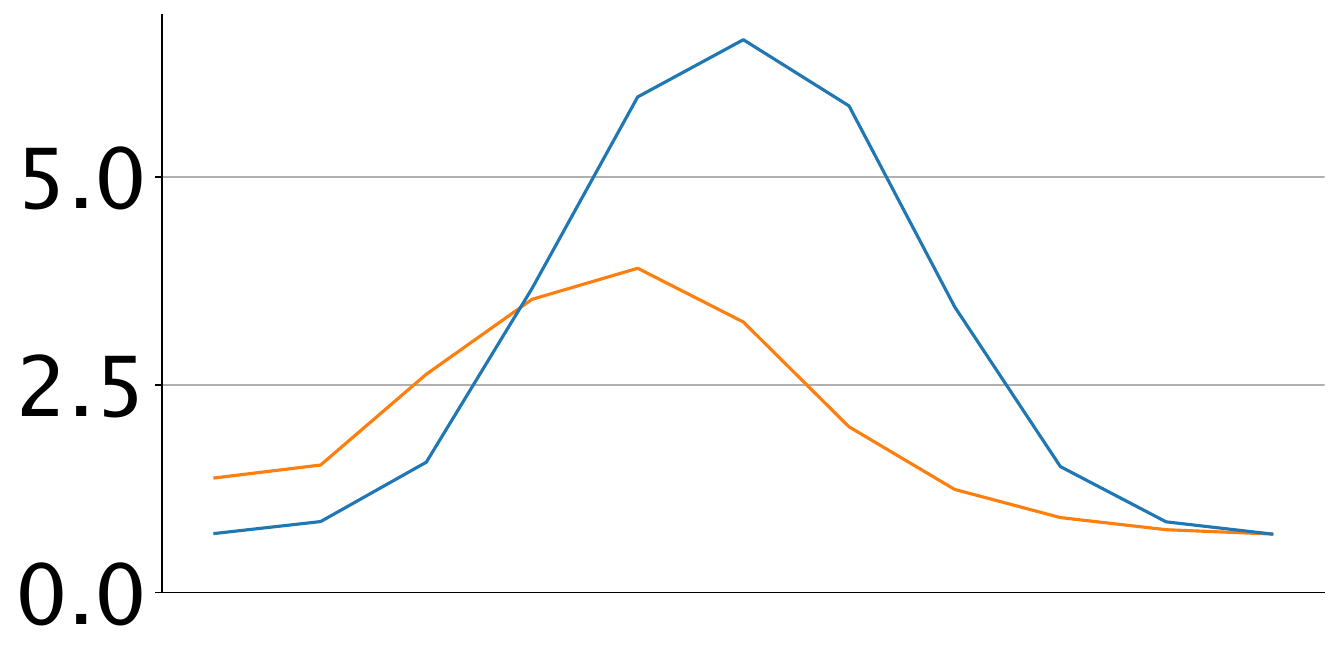} &
        \includegraphics[width=.24\linewidth]{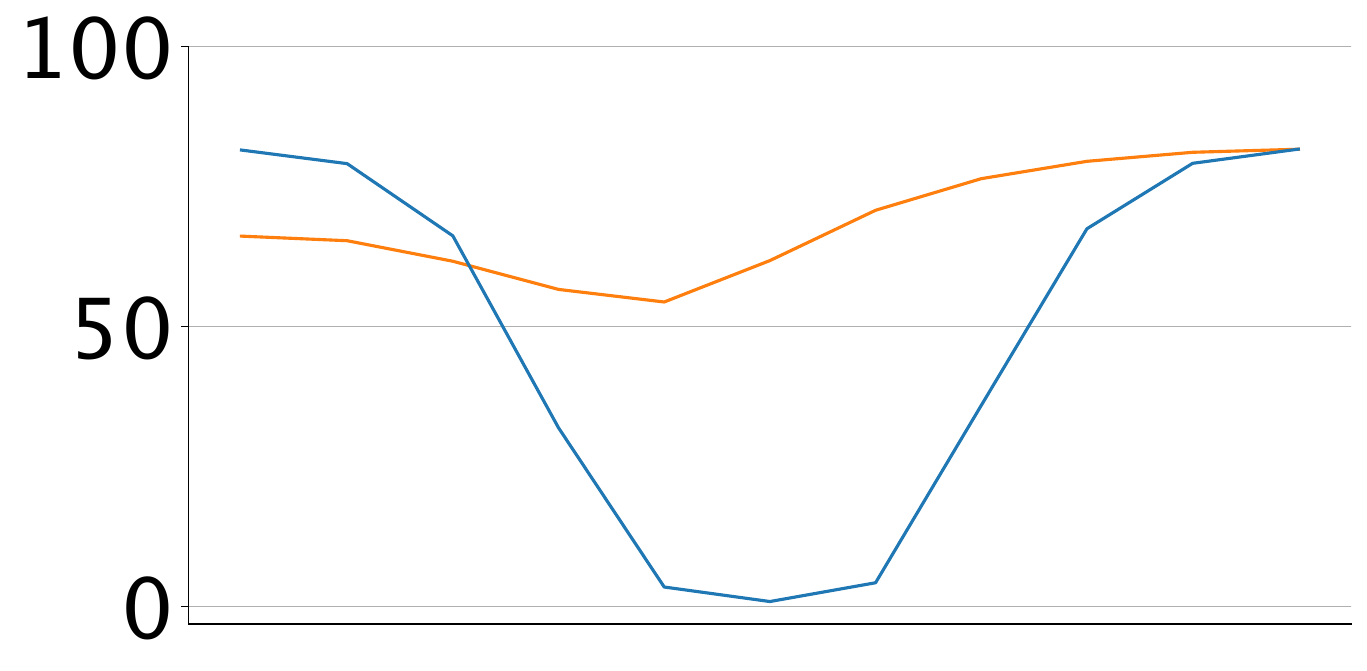} &
        \includegraphics[width=.24\linewidth]{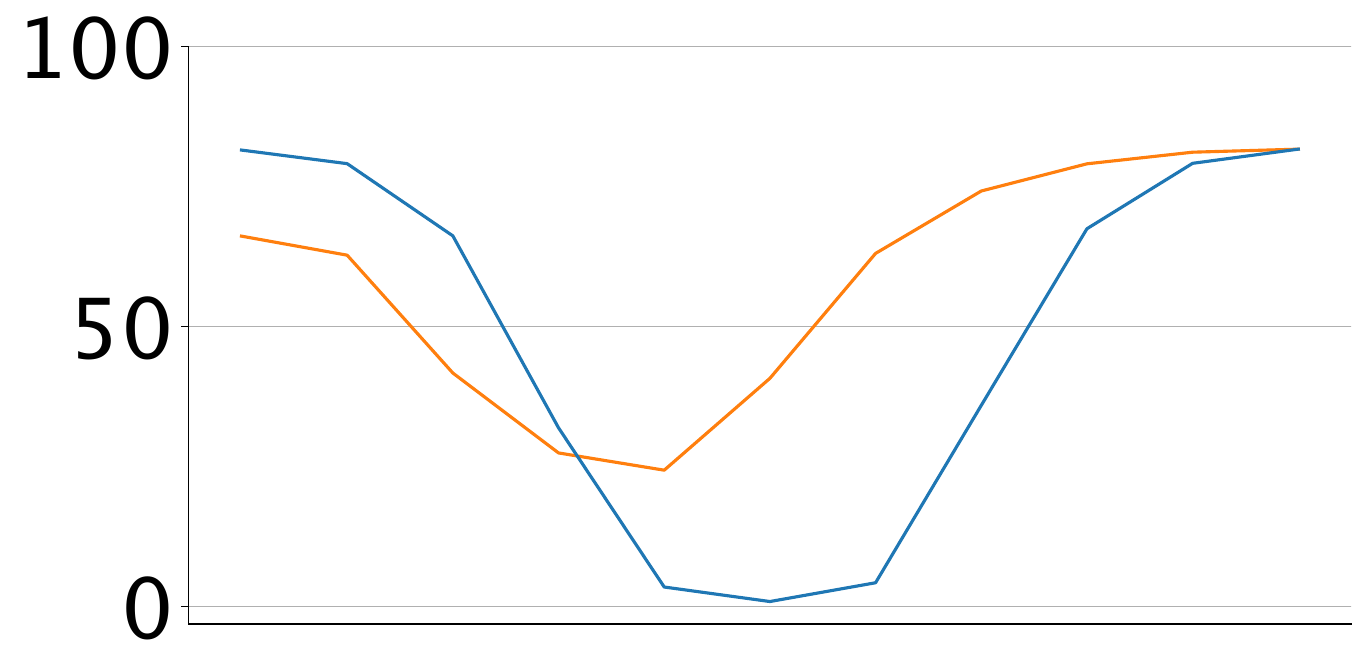}
        \\
        star $\theta^\star$ - source $Z$ & star $\theta^\star$ - heldout $H$ & 
        star $\theta^\star$ - source $Z$ & star $\theta^\star$ - heldout $H$ 
        \\
        \multicolumn{2}{c}{Train loss} & 
        \multicolumn{2}{c}{Train accuracy}
    \end{tabular}
    \caption{\textbf{Loss barriers for ResNet18 star models trained on ImageNet-1k}: training loss and accuracy curves obtained upon interpolation between \starcolor{star-regular} and \regcolor{regular-regular} model pairs. While ImageNet models struggle to achieve LMC, \starcolor{star-regular} barriers still fare much better than \regcolor{regular-regular} barriers.
    }
    \label{figure:resnet_imagenet_interpolation_plots}
\end{figure*}

Our main results are reported in \cref{figure:cifar_resnet_interpolation_plots} and \cref{table:star_model_empirical_verification}, in the main paper. For the sake of completeness, we report interpolation plots for the rest of our experiments in this section. We include plots for DenseNet (\cref{figure:densenet_cifar_interpolation_plots}), VGG (\cref{figure:vgg_cifar_interpolation_plots}), and ImageNet-1k (\cref{figure:resnet_imagenet_interpolation_plots}).

\section{Justifying the statement of the star domain conjecture}
\label{appendix:barrier_relationships}

While stating the star domain conjecture (\cref{conjecture:star_domain}), we specify a linear relationship, defined by the factor $\alpha$, between the minimum network widths required for our conjecture and the convexity conjecture to be valid. Here, we provide theoretical intuition toward this formulation.

\paragraph{Definitions and assumptions.}
For a given task, we define $B_r$ and $B_*$ as the average, over the posterior distribution, of the minimum achievable \quotes{regular-regular} and \quotes{star-regular} barriers, \ie

\begin{equation}
    B_r = \sum_{\theta_1, \theta_2 \sim \mathcal{P}}B_{min}(\theta_1, \theta_2)
\end{equation}

and 

\begin{equation}
    B_* = \sum_{\theta \sim \mathcal{P}, \theta^* \sim \mathcal{P}^*}B_{min}(\theta^*, \theta)
\end{equation}

where $B_{min}(\theta_1, \theta_2)$ is the lowest achievable barrier between two solutions $\theta_1, \theta_2$, and $\mathcal{P}, \mathcal{P^*}$ are the posterior distributions for the regular and star models, respectively. 
 
We make the following assumptions.

\begin{enumerate}

    \item $B_r$ and $B_*$ are functions of $w$, the network width, \ie $B_* = f_*(w)$ and $B_r = f_r(w)$, and $f_*$, $f_r$ are monotonically decreasing functions. This assumption is based on the observation that the wider the network, the lower the height of the loss barrier between two given solutions \cite{entezari_permutation_invariances_2022, ainsworth_git_re_basin_2022, ferbach2024proving}. We further assume \textbf{strict} monotonicity, so that $f_*, f_r$ are also invertible.

    \item Based on our empirical investigation in \cref{figure:cifar_wrn_width_depth}, we assume that $B_r$ and $B_*$ are linearly related, \ie 

    \begin{align}
    \label{eq:linear_relationship}
        B_* = f_*(w) = \alpha f_r(w) = \alpha B_r 
    \end{align}

    where $0 < \alpha < 1$.

    \item We define $w_r = f_r^{-1}(B_r)$ and $w_* = f_*^{-1}(B_*)$ as the minimum required network widths to achieve the \quotes{regular-regular} barrier $B_r$ and the \quotes{star-regular} barrier $B_*$, respectively. Based again on the empirically observed relationship  between network width and barrier, we further assume that $B_r$ and $w_r$ are inversely proportional to each other, \ie if $w_r = f_r^{-1}(B_r)$, then

        \begin{align}
        \label{eq:homogeneous_assumtion}
        f_r^{-1}(\alpha B_r) = f_r^{-1}(B_r) / \alpha =  w_r / \alpha
        \end{align}
        
\end{enumerate}. 

\paragraph{(Informal) proof.}

We aim to find widths $w_*$ and $w_r$ for which the two kinds of barriers are equal, \ie 

\begin{align}
    \label{eq:equating_barriers}
    f_*(w_*) = f_r(w_r)
\end{align}

But from \cref{eq:linear_relationship}, 

\begin{align}
    f_*(w_*) = \alpha f_r(w_*)
\end{align}

and therefore from \cref{eq:equating_barriers},

\begin{align}
    \alpha f_r(w_*) = f_r(w_r)
\end{align}

Applying $f_r^{-1}$ to both sides, 

\begin{align}
    \label{eq:inverting_barrier_function}
    f_r^{-1}(\alpha f_r(w_*)) = f_r^{-1}(f_r(w_r)) = w_r 
\end{align}

But from \cref{eq:homogeneous_assumtion}, $f_r^{-1}(\alpha f_r(w_*)) = f_r^{-1}(f_r(w_*))/\alpha$. By making this substitution to the left-hand side of \cref{eq:inverting_barrier_function}, we obtain $w_* / \alpha = w_r$, or $w_* = \alpha w_r$, \ie $w_*$ is a fraction of $w_r$, which is the result that we set out to prove.

While we empirically verify the assumptions used for obtaining this result, a more rigorous theoretical investigation may be carried out in future work.

\end{document}